\documentclass[reqno]{amsart}
\usepackage{graphicx}
\usepackage{amsmath,amssymb,amsthm} 
\usepackage{bm}
\usepackage{color}
\usepackage[scriptsize,tight]{subfigure}
\usepackage{mathabx}
\usepackage[font=small]{caption}
\usepackage{ctable}

\newcommand{\etalchar}[1]{$^{#1}$}
\newcommand{\bs}{\boldsymbol}

\DeclareMathOperator{\SE}{SE}

\DeclareMathOperator{\diff}{diff}

\newcommand{\RRR}{\mathbb{R}^3}
\newcommand{\T}{^{\top}}

\DeclareMathOperator{\divv}{div} %
\newcommand{\divs}{\divv_{S_{h}}}

\DeclareMathOperator{\id}{id}
\DeclareMathOperator{\diag}{diag}
\DeclareMathOperator{\shrink}{shrink}

\DeclareMathOperator{\ncc}{NCC}

\def \myfigspacer{0.1cm}
\newcommand{\cost}{\phi}
\newcommand{\otoprule}{\midrule[\heavyrulewidth]}

\title[Second-order Shape Optimization]{Second-order Shape Optimization for Geometric Inverse Problems in Vision}
\author{J. Balzer and S. Soatto}
\begin{document}

\maketitle

\begin{abstract}

We develop a method for optimization in shape spaces, i.e., sets of surfaces modulo re-parametrization. Unlike previously proposed gradient flows, we achieve superlinear convergence rates through a subtle approximation of the shape Hessian, which is generally hard to compute and suffers from a series of degeneracies. Our analysis highlights the role of mean curvature motion in comparison with first-order schemes: instead of surface area, our approach penalizes deformation, either by its Dirichlet energy or total variation. Latter regularizer sparks the development of an alternating direction method of multipliers on triangular meshes. Therein, a conjugate-gradients solver enables us to bypass formation of the Gaussian normal equations appearing in the course of the overall optimization. We combine all of the aforementioned ideas in a versatile \emph{geometric} variation-regularized Levenberg-Marquardt-type method applicable to a variety of shape functionals, depending on intrinsic properties of the surface such as normal field and curvature as well as its embedding into space. Promising experimental results are reported. 

\end{abstract}

\section{Introduction}

\subsection{Motivation}\label{subsec:motivation}

\begin{figure}[b]
\centering
\subfigure[]{\label{fig:teapot1}\includegraphics[width=0.45\columnwidth]{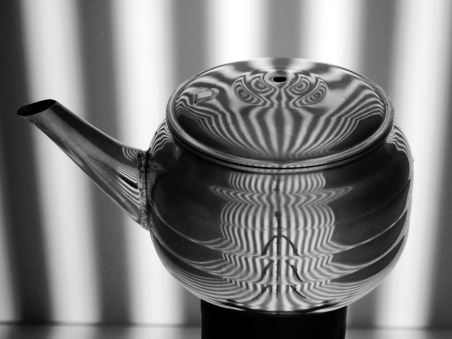}}\hspace{\myfigspacer}
\subfigure[]{\label{fig:teapot2}\includegraphics[width=0.45\columnwidth]{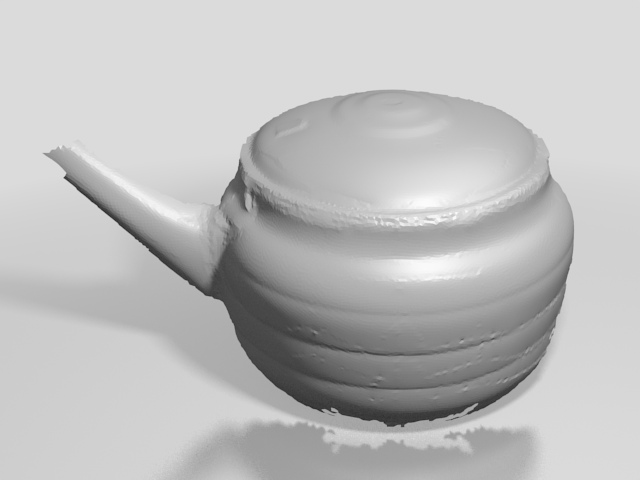}}
\caption{Application \emph{reconstruction of specular surfaces}: \subref{fig:teapot1} Correspondences between camera pixels and scene points viewed along the surface are established by a structured-light coding technique. \subref{fig:teapot2} The correspondences can be converted into normal information which is then integrated into a visible surface reconstruction by our method.}\label{fig:teapot}
\end{figure}

\begin{figure}[tb]
\centering
\subfigure[Noisy cube]{\label{fig:noisycube}\includegraphics[width=0.32\columnwidth]{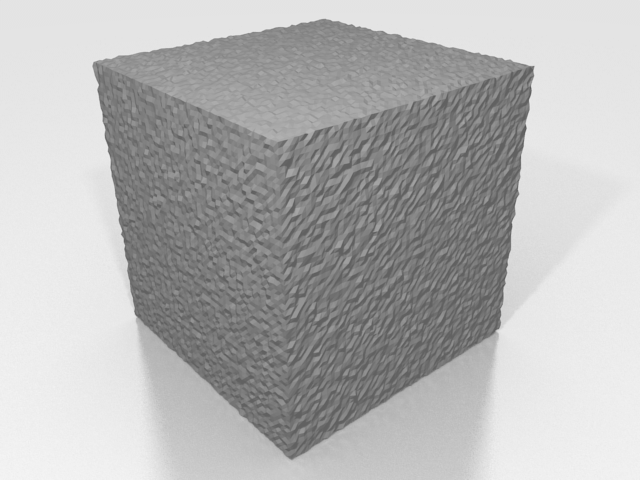}}\hfill
\subfigure[ML-LMTV denoising]{\label{fig:l2denoisednf}\includegraphics[width=0.32\columnwidth]{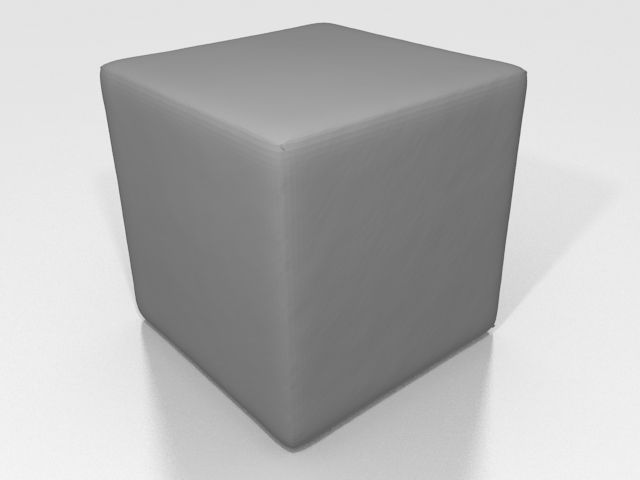}}\hfill
\subfigure[ROF-LMTV denoising]{\label{fig:l1denoisednf}\includegraphics[width=0.32\columnwidth]{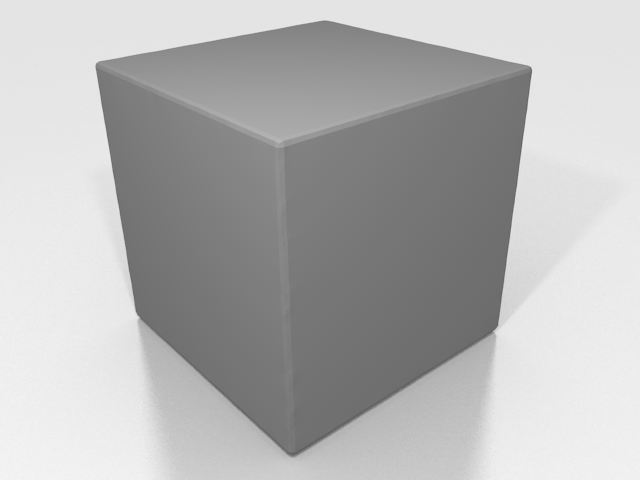}}
\caption{Application \emph{surface denoising}.}\label{fig:cubedenoising}
\end{figure}

\begin{figure}[tb]
\centering
\subfigure[Ground truth model]{\label{fig:mvsintro1}\includegraphics[width=0.32\columnwidth]{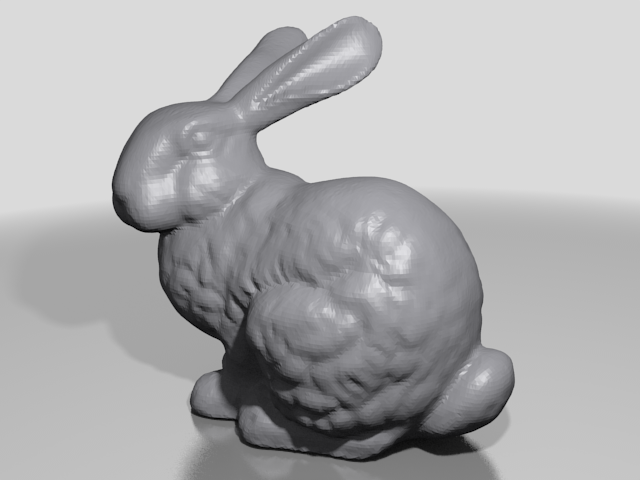}}\hfill
\subfigure[Textured model]{\label{fig:mvsintro2}\includegraphics[width=0.32\columnwidth]{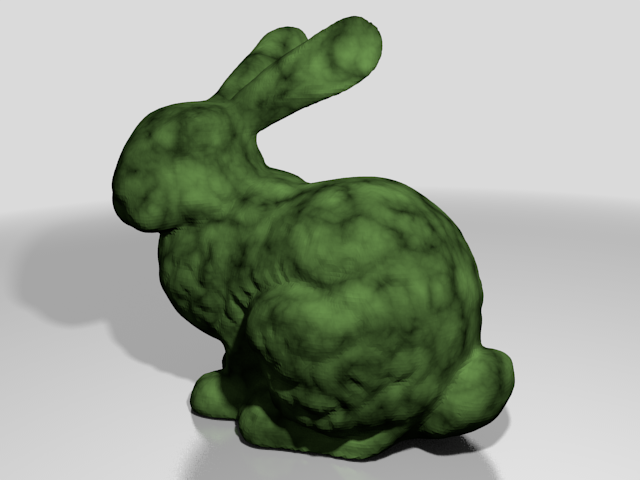}}\hfill
\subfigure[Image series]{\label{fig:mvsintro3}\includegraphics[width=0.32\columnwidth]{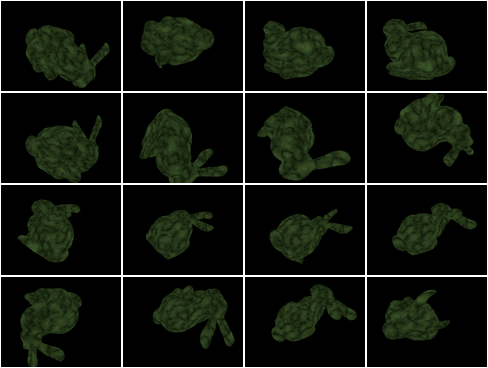}}
\caption{Application \emph{photometric optimization}.}\label{fig:mvsintro}
\end{figure}

Many inference tasks in vision amount to solving \emph{inverse problems}, where a solution is understood to be the element $x$ in a set $X$ which, given some model $f:X\times M\mapsto M$, minimizes the \emph{residual} $r(x)=f(x,s)-t$ between a \emph{signal} $s$ and its prediction under $f$. For instance, in optical flow, one wishes to minimize the distance between an image $\mathcal{I}_s$ and a warped version $\mathcal{I}_t\circ w$ of $\mathcal{I}_s$ w.r.t. $w$ in the diffeomorphism group $X=\diff(D)$ of the image plane. In this paper, we are interested in the case where $M$ is a linear space of functions, e.g.,   $BV(D)$ or $H^1(D)$, over some geometric domain $D$, but -- quite similar to the example of $\diff(D)$ -- the set of latent variables $X$ is \emph{not}, but instead a {\em shape space}, consisting of three-dimensional (3-d) surfaces up to re-parametrizations. The literature offers a wealth of first-order numerical methods. But despite their superior convergence properties, {\em to this date there are no generally applicable second-order methods} for shape optimization. This is explained by the difficulties in accurately and efficiently approximating the Hessian. We focus on a class of separable quadratic functionals to propose what is, to the best of our knowledge, {\em the first second-order numerical method for solving visual inference problems on shape spaces.} This is our first contribution. As shown in Sect.~\ref{subsubsec:shapelm}, the construction suppresses eigenspaces of the Hessian which are responsible for shrinking biases in traditional gradient flows. To ensure regularization, we suggest penalizing variations, not of the iterated surface itself, but \emph{deformations} thereof. This leads to a variant of the classic Levenberg-Marquardt method which can be applied under weak assumptions on $f$ by breaking down the nonlinear and possibly nonconvex global optimization problem into a sequence of convex subproblems. Depending on the choice of regularizer, one type of subproblem encountered is equivalent to the Rudin-Osher-Fatemi (ROF) model for image denoising and segmentation~\cite{Rudin1992}. To solve it numerically, we develop an extension of the alternating direction method of multipliers (ADMM, a.k.a. Split Bregman~\cite{Goldstein2009}) to surfaces represented by triangular meshes~(Sect.~\ref{subsec:sbcgls}). This is our second contribution. We demonstrate that the chosen class of separable quadratic functions applies to a variety of problems relevant to vision, from mesh reconstruction from unorganized point clouds and deflectometric images (Fig.~\ref{fig:teapot}), surface denoising (Fig.~\ref{fig:cubedenoising}), to photometric optimization (Fig.~\ref{fig:mvsintro}), which will all be explored in Sect.~\ref{sec:casestudies}. Finally, we plan to distribute the code implementing each application upon completion of the anonymous review process.

\subsection{Relation to prior work}

The natural question arises why we should develop \emph{local} methods -- even of second order -- when globally optimizable convex programs for many reconstruction tasks have been proposed, cf.~\cite{Boykov2006,Kolev2009}. The short answer is that first, in these models, convexity originates from embedding the unknown surface into a linear space through some implicit representation such as a distance or characteristic function. We would like to avoid such resource-hungry representations as much as possible and restrict their use to as-coarse-as-possible initialization. Second, as soon as visibility, which in turn depends on the optimization variable itself, is fully considered in these models, convexity  will be lost. There are some analogies between the present paper and~\cite{Ochs2013} in the sense that the problem of interest is decomposed into a sequence of nondifferentiable subproblems: The latter generalizes Cand\'es' reweighted $\ell_1$-algorithm, and the goal of the decomposition is to handle nonconvex regularizers. Sect.~\ref{subsubsec:shapelm} is an extension of~\cite{Balzer2012}, where a regularization-free \emph{Gauss-Newton} method was presented especially for normal field integration, to a much wider class of cost functions. The ADMM has been adapted to linear spaces over surfaces before, first by Wu et al.~\cite{Wu2011}, later by Liu and Leung~\cite{Liu2012}. The authors of the latter paper are concerned with point-based surface models. The former approach is different from ours in that it explicitly forms Gaussian normal equations at every iteration. In comparison, we suggest executing a few preconditioned conjugate-gradient steps on the corresponding \emph{overdetermined} linear system. A similar trick has been proposed previously for large-scale bundle adjustment~\cite{Byrod2010}. The optimization framework developed in this paper is fairly general but applied to the sample problems in Sect.~\ref{sec:casestudies}, it inherits some of the ideas found in the specialized literature: Similar to Avron et al.~\cite{Avron2010}, we couple denoising of the normal field with subsequent integration for the purpose of surface fairing (Sect.~\ref{subsec:denoising}) and reconstruction (Sect.~\ref{subsec:pclreconstruction}).  Inspired by~\cite{Kazhdan2013}, we consider orientation information for reconstruction but prefer explicit surface models and account for the nonlinearity of $f$, the Gauss map. Geometric applications of the Split Bregman method have been studied in~\cite{Goldstein2009}, but different from Sect.~\ref{subsec:pclreconstruction} within a level set segmentation framework. The body of literature on our third sample  application -- photometric optimization from multiview stereo images -- is too vast to do it justice here. Let us only explicitly mention the works~\cite{Delaunoy2010,Jin2005,Tylecek2010} because they feature shape optimization albeit of first order only.

\section{Main result}

\subsection{From Gauss-Newton to Levenberg-Marquardt}\label{subsec:lm}

With the notation introduced at the beginning of Sect.~\ref{subsec:motivation}, any nonlinear least-squares (LS) problem takes the form 
\begin{equation}\label{eq:problem}
	\min\limits_{x\in X} \frac{1}{2}\|r(x)\|^2_{L^2(D)}.
\end{equation}
The default optimization strategy is the Gauss-Newton algorithm, which exploits the fact that the Hessian of the $L^2$-energy at some $x_k\in X$ can be accurately approximated by the differential $Dr$ of the residual: $D^2E_d|_{x_k}\approx (Dr|_{x_k})\T Dr|_{x_k}$ in $x_k$. In combination with the classic Newton method, this gives the implicit time step
\begin{subequations}\label{eq:gaussnewton}
\begin{equation}\label{eq:gaussnewton1}
  x_{k+1}= x_k + v
\end{equation}
in which the \emph{update} $v$ solves the linear equation system
\begin{equation}\label{eq:gaussnewton2}
 Dr\T Dr v=-Dr\T r.
\end{equation}
\end{subequations}
An illustrative interpretation of~\eqref{eq:gaussnewton} is the following: Approximating $r$ by its first order Taylor expansion $r(v)\approx r(x_k)+Dr|_{x_k}v$ results in a \emph{local linear} LS problem 
\begin{equation}\label{eq:localleastsquares}
	\min_v\frac{1}{2}\|r(x_k)+Dr|_{x_k}v\|^2_{L^2(D)}
\end{equation}
whose normal equation is precisely~\eqref{eq:gaussnewton2}. Returning to the introductory example of optical flow, the Gauss-Newton method applied to the \emph{nonlinear} gray value conservation law under a translational deformation model, in which $w$ is assumed spatially constant, is equivalent to the Lucas-Kanade algorithm~\cite{Baker2004} (apart from the multiscale strategy it includes to avoid falling into local minima). 

Occasionally, Eq.~\eqref{eq:gaussnewton2} becomes underconstrained which causes ill-conditioning of $Dr\T Dr$ by creating zeros among its eigenvalues. In the Levenberg-Marquardt method in its original form, the issue is addressed by adding $\frac{\lambda}{2}\|v\|^2_{L^2(D)}$  to~\eqref{eq:localleastsquares}. The new local approximation of the energy limits the step size (i.e., the size of the \emph{trust region}) inversely proportional to the magnitude of $\lambda$. The regularizer appears in the normal matrix as $\lambda$-fold multiple of the identity, and hence, pushes the spectrum of the former towards positive values by an amount of $\lambda$. When the update step $v$ possesses some form of spatial regularity, we can punish large \emph{variations} in lieu of large magnitudes of $v$ by replacing~\eqref{eq:localleastsquares} with
\begin{equation}\label{eq:localleastsquaresreg}
	\min_v\frac{1}{2}\|r(x_k)+Dr|_{x_k}v\|^2_{L^2(D)}+\frac{\lambda}{p} \|Dv\|^p_{L^p(D)}.
\end{equation}
For $p=2$, the regularizer equals the Dirichlet energy which maintains linearity of the local LS problem. For $p=1$, Eq.~\eqref{eq:localleastsquaresreg} becomes the ROF functional. Bachmayr and Burger point out this connection in~\cite{Bachmayr2009}. \emph{The motivation of the present paper is to generalize the resulting variation-regularized Levenberg-Marquardt method consisting of Eqs.~\eqref{eq:gaussnewton1} and \eqref{eq:localleastsquaresreg} from vector spaces to sets of surfaces in 3-d.} The necessary theoretical foundations are laid out in the following section before we state our main result in Sect.~\ref{subsubsec:shapelm}.

\subsection{Formulation in shape space}\label{subsec:shapeoptim}

\subsubsection{Shape spaces, functions, and gradient flows}\label{subsubsec:gradientflows}

When we speak of \emph{shape}, we mean the trace of a surface, i.e., the collection of its points in a set-theoretic sense modulo its symmetry group, which consists of all smoothness-preserving re-parametrizations. Let $S_0$ be the boundary of a smooth reference subdomain of $\mathbb{R}^3$. The set of all diffeomorphic embeddings $\diff(S_0,\mathbb{R}^3)$ becomes a \emph{shape space} as soon as two embeddings $w,z\in\diff(S_0,\mathbb{R}^3)$ are considered equivalent if they differ by some $\tau\in\diff(S_0,S_0)$, i.e., $w=z\circ \tau$. Let us remark that in particular, all elements of $X=\diff(S_0,\mathbb{R}^3)/\diff(S_0,S_0)$ exhibit the same topology, namely that of $S_0$. A shape space has the structure of an infinite-dimensional manifold~\cite{Delfour2001}. We will not worry about its intriguing geometrical and topological properties. For all practical purposes, it suffices to acknowledge that the tangent space of this manifold at a ``point'' $S$ consists of all infinitesimal normal velocities $v$ in $H^1(S)$ respectively $BV(S)$. This is quite intuitive: tangential deformations map surface points to surface points, do not alter shape, and hence preserve the equivalence class of $S$. We can also conduct analysis. An important example of a shape function is the surface integral
\begin{equation}\label{eq:shapefunctional}
	E(S)=\int\limits_S \cost(S) \,\mathrm{d}S.
\end{equation}
We admit costs $\cost(\bm{x},\bm{n})$ depending on $\bm{x}\in\mathbb{R}^3$ as well as the unit surface normal $\bm{n}\in \mathbb{S}^2$ but generalizing what follows to higher-order differential surface properties (e.g, the Willmore energy, cf. Appx.~\ref{appx:curvature}) is possible. The \emph{shape differential} of $E$ at some $S$ in the tangential direction of $v$ is given by
\begin{equation}\label{eq:shapedifferential}
	DE(S;v)=\int\limits_S \underbrace{\left(\kappa\cost+\langle\nabla\cost,\bm{n}\rangle-\nabla\T_S\nabla_{\mathbb{S}^2}\cost\right)}_{g_E} v \,\mathrm{d}S,
\end{equation}
where $\kappa$ denotes the mean curvature and $g_E$ the \emph{shape gradient} of $E$. A derivation of this formula can be found in several places, cf.~\cite{Delfour2001,Goldluecke2007,Jin2005,Solem2005b}, its application in many more, cf.~\cite{Chang2007,Delaunoy2010,Tylecek2010}. Note that the domain of $\cost$ may extend to the embedding space $\mathbb{R}^3\times\mathbb{R}^3 \supset S\times \mathbb{S}^2$. Correspondingly, $\nabla$ is the \emph{Euclidean} nabla operator, whereas $\nabla_S$ and $\nabla_{\mathbb{S}^2}$ denote the \emph{intrinsic} or \emph{surface gradient} on $S$ respectively the unit sphere $\mathbb{S}^2$. Also note that because $\ker\nabla\T_S=[\bm{n}(S)]$, it is sufficient to calculate the Euclidean derivative of $\cost$ w.r.t. $\bm{n}$ without reprojecting onto $\mathbb{S}^2$. By evolving some $S_0$ in the steepest descent direction $-g_E$ according to
\begin{equation}\label{eq:gradientdescent}
	S_{k+1}=S_k-(\kappa\cost+\langle\nabla\cost,\bm{n}\rangle-\nabla\T_S\nabla_{\mathbb{S}^2}\cost)\bm{n}(S_k),
\end{equation}
we can decrease~\eqref{eq:shapefunctional} in two ways: either by reducing the surface area\footnote{The area integral measures -- up to some material properties inherent in $\cost$ -- the \emph{tangential strain} or \emph{membrane energy} of a surface.} via (weighted) mean curvature motion (MCM) in the direction $-\kappa\cost$; alternatively, we let each point follow the direction of greatest decrease of the cost function $\nabla\cost$ respectively $\nabla_{\mathbb{S}^2}\cost$. The stationary point of the descent, at which $g_E=0$, will be determined by the equilibrium between these two forces. This equilibrium is responsible for a phenomenon called \emph{minimal surface bias}: First, whenever the descent direction w.r.t. $\cost$ is uninformative in the sense that  $\nabla\cost=\bm{0}$ while $\cost>0$, the evolution will locally favor surfaces of minimal area. Second, due to the counterforce, the limit surface cannot fully account for the regularity of $\cost$ leading to visible oversmoothing and retraction of boundaries if present. Finally, where both $\cost$ \emph{and} its derivatives w.r.t. $\bm{x}$ and $\bm{n}$ vanish\footnote{As an example, consider the re-projection error of multiple views onto a homogeneously textured surface region.}, the evolution~\eqref{eq:gradientdescent} will stagnate. 

\subsubsection{Hessian-free superlinear optimization}\label{subsubsec:shapelm}

While the computation of first-order shape differentials like~\eqref{eq:shapedifferential} is relatively straightforward, nonzero geodesic curvature of shape spaces significantly aggravates this process for second-order derivatives~\cite{Delfour2001}. So far, the lack of symmetry and positive-definiteness have defeated any attempt to implement a pure Newton method for~\eqref{eq:shapefunctional}. Our key insight is that this problem can be circumvented under the condition that $\cost$ is \emph{separable} and \emph{quadratic}:
\begin{equation}\label{eq:separability}
\cost(\bm{x},\bm{n})=\frac{1}{2}(\|\bm{r}_x(\bm{x})\|^2+\|\bm{r}_n(\bm{n})\|^2).
\end{equation}
The residual $\bm{r}_x$ of $\cost$ over $\mathbb{R}^3$ arises from the (dis)location of surface points in space. Note that $\bm{r}_x$ is generally vector-valued, e.g., to account for multi-channel images or distances to known points (Sect.~\ref{subsec:pclreconstruction} and~\ref{subsec:mvs}). The shape differential of $\bm{r}_x$, describing the impact of infinitesimal normal deformations $v$ on the value of $\bm{r}_x$, is directly given by $D\bm{r}_x(v\bm{n})$. In perfect analogy, the normal error $\bm{r}_n$ is a map taking $\bm{n}\in\mathbb{S}^2$ to the embedding space\footnote{The example of the difference between two unit vectors shows that clearly the image of $\bm{r}_n$ is not necessarily contained in $\mathbb{S}^2$.} $\mathbb{R}^3$ with Jacobian $D_{\mathbb{S}^2}\bm{r}_n: T\mathbb{S}^2\to T\mathbb{R}^3$. Invocation of the chain rule yields $-D_{\mathbb{S}^2}\bm{r}_n\nabla_S v$ for the shape differential of $\bm{r}_n$. Here, we have used the fact that pure infinitesimal rotations of the normal are related to the velocity $v$ by its negative surface gradient $-\nabla_S v$, cf.~\cite[Prop. 1]{Balzer2012}. The shape differentials of $\bm{r}_x$ and $\bm{r}_n$ enable a local quadratic approximation
\begin{equation}\label{eq:localshapefunction}
	E_d(v):=\frac{1}{2}\|\bm{r}_x(\bm{x})+D\bm{r}_x(v\bm{n})\|^2_{L^2(S)}\\+\frac{1}{2}\|\bm{r}_n(\bm{n})-D_{\mathbb{S}^2}\bm{r}_n\nabla_S v\|^2_{L^2(S)}
\end{equation}
of~\eqref{eq:shapefunctional} around $S$. The equivalence of~\eqref{eq:localleastsquaresreg} and~\eqref{eq:gaussnewton2} then immediately implies a shape space analogue of~\eqref{eq:gaussnewton}:
\begin{subequations}\label{eq:shapegaussnewton}
\begin{equation}\label{eq:shapegaussnewton1}
  S_{k+1}= S_k + v\bm{n}(S_k)
\end{equation}
where the normal velocity $v$ is the unique minimizer of  
\begin{equation}\label{eq:shapegaussnewton2}
 	E_d(v)+\frac{\lambda}{p}\|\nabla_{S}v\|^p_{L^p(S_k)}. 
\end{equation}
\end{subequations}
As shown in Sect.~\ref{subsec:sbcgls}, there are efficient ways of minimizing this function (for fixed $S_k$ and $p=1,2$). 

Remarkably, while the steepest descent~\eqref{eq:gradientdescent} used in previous approaches strives to reduce surface area, the solution of the local subproblem~\eqref{eq:shapegaussnewton2} does not. The simple explanation is that minimization is performed w.r.t. the velocity field $v$ and coupled with the properties of the surface only through the shape differentials of $\bm{r}_x$ and $\bm{r}_n$. Unfortunately, problems arise if the cost $\cost=\cost(\bm{x})$ is independent of the normal, like in the applications discussed in Sects.~\ref{subsec:pclreconstruction} and~\ref{subsec:mvs}. When $\bm{r}_n=\bm{0}$, the minimizer of~\eqref{eq:localshapefunction} can be obtained in closed form: 
\[
	v\bm{n}=(D\bm{r}_x)^{-1}\bm{r}_x.
\]
This, however, requires $D\bm{r}_x$ to be of full rank, a condition which can never hold in the vicinity of a stationary point where $D\bm{r}_x$ should be identically zero. A more intuitive explanation is the following: Loss of the mean curvature term in the descent rule cannot remain without consequences. Surface area correlates with surface smoothness. Without the binding influence of $\kappa$, points on the surface will be able to move around \emph{separately}, quickly compromising its integrity unless $\bm{r}_x$ is unrealistically smooth. The regularization term in~\eqref{eq:shapegaussnewton2} comes to the rescue by enforcing either harmonic ($p=2$) or piecewise constant ($p=1$)  \emph{descent directions} or \emph{deformations} in~\eqref{eq:shapegaussnewton1}. We can only conjecture that the latter do not favor smooth surfaces. A convenient side effect is that the regularizer will inpaint nonzero values of $v$ to regions where both $\bm{r}_x$ and $D\bm{r}_x$ vanish and a gradient descent would come to a complete halt (as discussed at the end of Sect.~\ref{subsubsec:gradientflows}). 

\begin{figure}
\subfigure[]{\label{fig:imagedenoising1}\includegraphics[width=0.45\columnwidth]{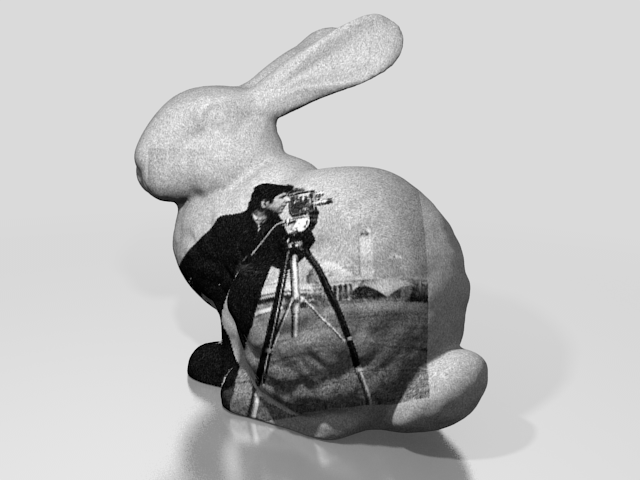}\hspace{0.01\columnwidth}\hspace{\myfigspacer}\includegraphics[width=0.45\columnwidth]{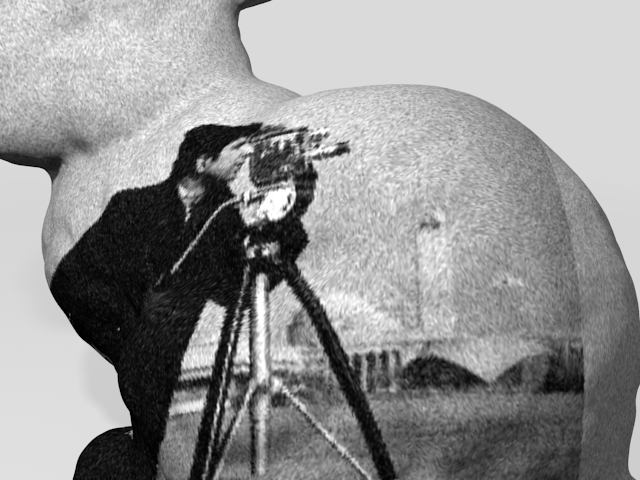}}\hspace{\myfigspacer}
\subfigure[$\lambda=7$]{\label{fig:imagedenoising3}\includegraphics[width=0.45\columnwidth]{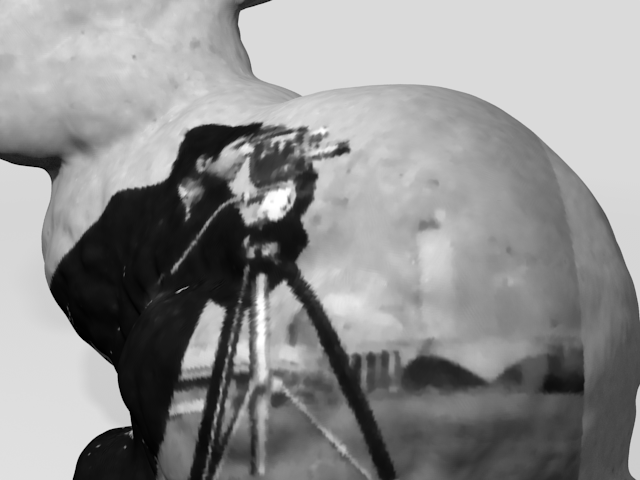}}\hspace{\myfigspacer}
\subfigure[$\lambda=10$]{\label{fig:imagedenoising4}\includegraphics[width=0.45\columnwidth]{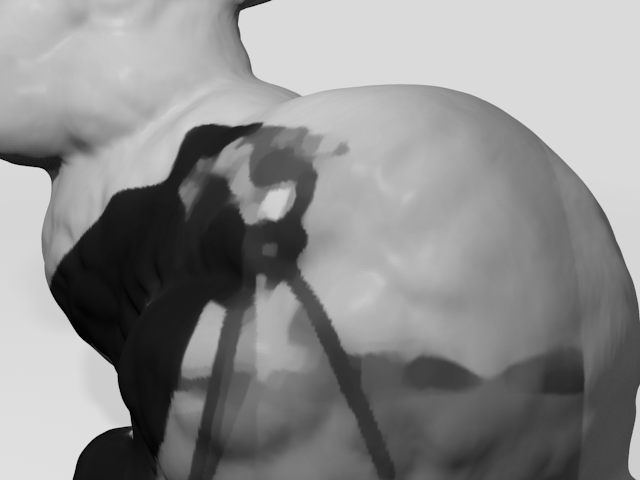}}\\
\caption{\subref{fig:imagedenoising1} The cameraman test image with additive Gaussian noise of standard deviation $\sigma=20$ texture-mapped onto the Stanford bunny. \subref{fig:imagedenoising3}-\subref{fig:imagedenoising4} Results of ROF denoising on the surface with different regularization weights.}
\label{fig:imagedenoising}
\end{figure}

\subsection{Conjugate gradient ADMM on triangular meshes}\label{subsec:sbcgls}

To minimize~\eqref{eq:shapegaussnewton2} efficiently, we now describe a variation of the ADMM on surfaces. Since, in the end, we are interested in designing a numerical algorithm, let us consider finite-dimensional representations of $S$ and the function spaces on it. In particular, let us assume we have a triangulation $S_h $ of $S$ but emphasize that the continuous formulation in Sect.~\ref{subsubsec:shapelm} equally admits other kinds of discretizations, like e.g. with zero-sets of a scalar-valued function on $\mathbb{R}^3$. The precise details, in particular the lengthy derivation of the mass matrices $\mathbf{W}_x$ and $\mathbf{W}_n$ as well as the discrete nabla operator $\nabla_{S_h}$ on $S_h$, is deferred to Appx.~\ref{appx:fem}. We collect the Jacobians and residuals from~\eqref{eq:localshapefunction} in
\[
	\mathbf{A}=\left(\begin{array}{c} \mathbf{W}_x\diag(D\bm{r}_x) \\ -\mathbf{W}_n\diag (D_{\mathbb{S}^2}\bm{r}_n)\nabla_{S_h}\end{array}\right),\;\;\mathbf{f}=\left(\begin{array}{c}\mathbf{W}_x\mathbf{r}_x\\\mathbf{W}_n\mathbf{r}_n\end{array}\right). 
\]
The upper block-diagonal matrix is assembled from the values that the corresponding continuous quantities take at the vertices, the lower half respectively from the values on the faces. With these abbreviations in place, starting from $\mathbf{v}_0=\mathbf{d}_0=\mathbf{b}_0=\mathbf{0}$, the ADMM for minimization of~\eqref{eq:shapegaussnewton2} iterates the following three steps:
\begin{subequations}\label{eq:sb}
\begin{align}\label{eq:sb1}(\mathbf{A}^{\top}\mathbf{A}-\lambda\mu\nabla_{S_h}^{\top}\nabla_{S_h})\mathbf{v}_{k+1}=\mathbf{A}^{\top}\mathbf{f}+\lambda\mu\nabla_{S_h}^{\top}\mathbf{d}_k,\\\label{eq:sb2}
\mathbf{d}_{k+1}=\shrink(\nabla_{S_h} \mathbf{v}_{k+1}+\mathbf{b},\mu^{-1}),\\\label{eq:sb3}
\mathbf{b}_{k+1}=\mathbf{b}_k+\nabla_{S_h} \mathbf{v}_{k+1}-\mathbf{d}_{k+1}.
\end{align}
\end{subequations}
\begin{table}[tb]
\centering
 \setlength{\tabcolsep}{4pt}
\def \mycolwidth{0.55cm}
\scriptsize
\begin{tabular}{lcccccc}\toprule
 & Cube ML & Cube ROF & Teapot & Sphere  & MVS  \\\otoprule %
$n$ & $24,578$  & $24,578$ & $17,974$ & $7,842$ & $34,834$\\\otoprule
GD & $0.94$ & $0.89$ & $0.56$ & $0.43$ &  $30.9$  \\\midrule
LMD  & $0.68$ & $0.67$ & $1.98$ & $0.48$ &  $28.4$ \\\midrule
LMTV  & $1.52$ & $1.5$ & $7.2$ & $1.41$ &  $36.8$ \\\bottomrule
\end{tabular}
\caption{Execution time for a single step in seconds. The first row contains the number $n$ of vertices in the optimized mesh.}
\label{tab:steptimes}
\end{table} 
Note that the roles of $\mu$ and $\lambda$ have switched opposed to the canonical notation in~\cite{Goldstein2009}. The reason is that our focus is on the Levenberg-Marquardt method here, in which the parameter discounting the step length is conventionally referred to by $\lambda$. We make the following modification to the original algorithm and its surface-based variant proposed in~\cite{Wu2011}: First, note that~\eqref{eq:sb1} is the Gaussian normal equation of the LS problem associated with~\eqref{eq:localshapefunction}. The only advantage of working with the normal equation is that therein, $\mathbf{A}\T\mathbf{A}$ becomes symmetric and strictly diagonally-dominant. This is exploited in~\cite{Goldstein2009} by invoking a simple and very efficient Jacobi solver. At the same time, small eigenvalues will become even smaller with deteriorating influence on the condition number. Additionally, discrete divergence and Laplace-Beltrami operators defined by $\divs:=\nabla_{S_h}\T$ respectively $\Delta_{S_h}:=\nabla_{S_h}\T\nabla_{S_h}$ are inconsistent with discrete conservation laws, which may lead to numerical instabilities~\cite{Desbrun2005}. Last but not least, there is the cost of computing the matrix product. Therefore, we propose to rearrange the normal equation of~\eqref{eq:localshapefunction} as follows:
\[
\left(\begin{array}{c} \mathbf{A}\\ \lambda\mu\nabla_{S_{h}}\end{array}\right)\mathbf{v}_{k+1}=\left(\begin{array}{c} \mathbf{f}\\ \lambda\mu \mathbf{d}_k\end{array}\right).
\]
This linear system is now overdetermined but amenable to the Conjugate Gradients Least-Squares method~\cite{Bjorck1996}, which avoids explicit formation of the normal equation. Its iterative nature allows us to preserve the inexactness of the original ADMM. When $p=2$ and hence~\eqref{eq:localshapefunction} is differentiable, setting  $\mu=1$ and $\mathbf{d}_k=\mathbf{0}$, we immediately obtain the update $\mathbf{v}_{k+1}$ from~\eqref{eq:sb1} without the need for shrinkage~\eqref{eq:sb2} and executing Bregman steps~\eqref{eq:sb3}. 

\begin{figure}[tb]
\centering
\subfigure[GD]{\label{fig:teapotgrad}\includegraphics[width=0.45\columnwidth]{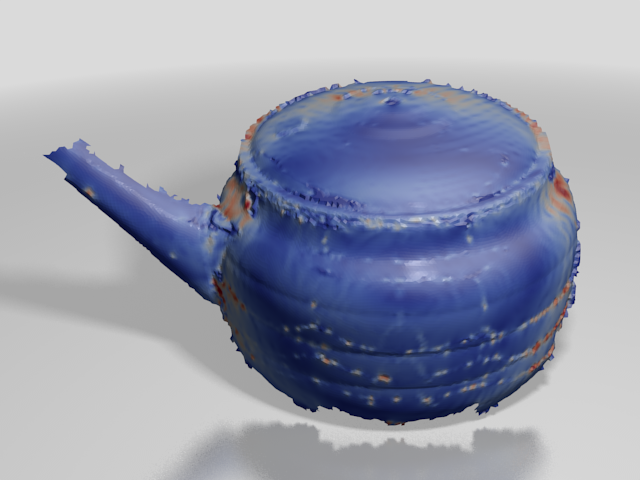}}\hspace{\myfigspacer}
\subfigure[Our method]{\label{fig:teapotlm}\includegraphics[width=0.45\columnwidth]{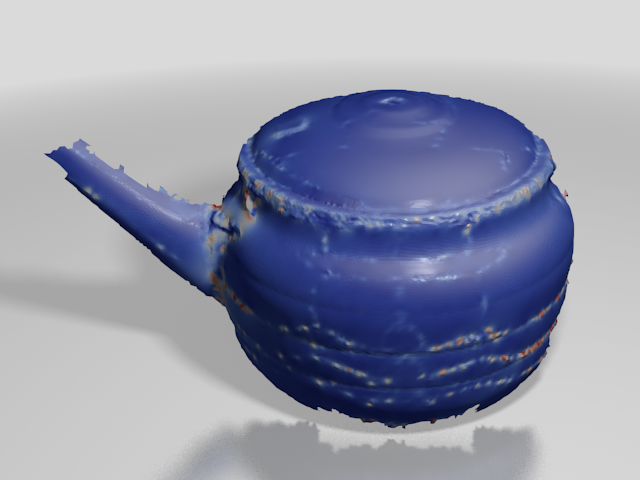}}
\caption{Result from Fig.~\ref{fig:teapot2} overlayed with the colormapped error distribution in the interval $[0,0.5]$.} \label{fig:teapoterror}
\end{figure}

\section{Applications}\label{sec:casestudies}

All algorithms discussed in the following section were implemented in C++ and executed on a single 3.4 $\mathrm{GHz}$ core of a commodity computer with 12 $\mathrm{GB}$ of main memory. We will make all source code publicly available. To begin with, we showcase the viability of our ADMM variant at hand of texture denoising (Fig.~\ref{fig:imagedenoising}). Here, the surface remains static so we achieve essentially the same as~\cite{Wu2011,Liu2012}. Applications of the method introduced in Sect.~\ref{subsec:shapeoptim}, in which the surface itself plays the role of the optimization variable, will be presented in the following sections. Thereby, we abbreviate the Levenberg-Marquardt method with a TV-regularizer ($p=1$) by LMTV respectively LMD when~\eqref{eq:shapegaussnewton2} contains the Dirichlet energy ($p=2$). We compare LMTV and LMD with the existing gradient descent (GD) scheme. Let us remark that its ad-hoc formulation~\eqref{eq:gradientdescent} does not directly lend itself to implementation because it suffers from numerical stiffness due to the MCM term. Noticing that $\kappa\bm{n}=\Delta_S(S)$, i.e., the mean curvature \emph{vector} $\kappa\bm{n}$ is just the Laplace-Beltrami operator $\Delta_S=\nabla\T_S\nabla_S$ applied to the functions that embeds $S$ into $\mathbb{R}^3$, we arrive at the backward Euler scheme
\begin{equation}\label{eq:impliciteuler}
	S_{k+1}+(\lambda+\cost)\Delta_S(S_{k+1})\\=S_{k}-(\langle\nabla\cost,\bm{n}\rangle-\nabla\T_S\nabla_{\mathbb{S}^2}\cost)\bm{n}.
\end{equation}
Additionally, a regularization weight $\lambda$ has been introduced as a factor of $\kappa$, amplifying the smoothing effect of MCM if necessary. The price to pay for stability is the inversion of the matrix $\id+(\lambda+\cost)\Delta_S$ at each iteration. Consequently, the number of floating point operations per gradient step is not significantly smaller than for each iteration in LMD, see Tab.~\ref{tab:steptimes}.

\begin{figure}[tb]
\centering
\subfigure[Noisy normal field]{\label{fig:denoising1}\includegraphics[width=0.32\columnwidth]{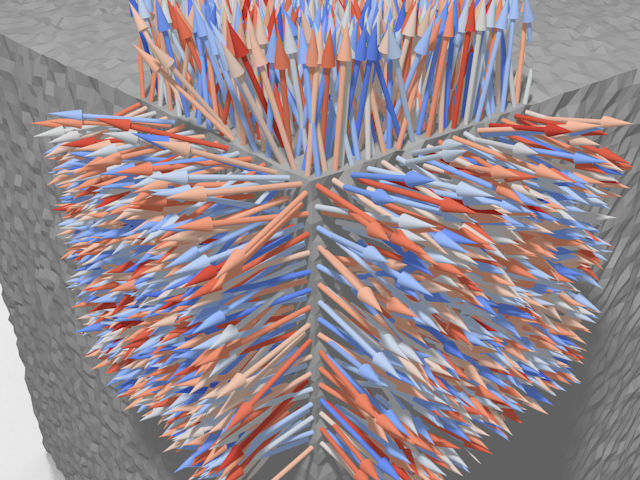}}
\subfigure[ML denoising]{\label{fig:denoising2}\includegraphics[width=0.32\columnwidth]{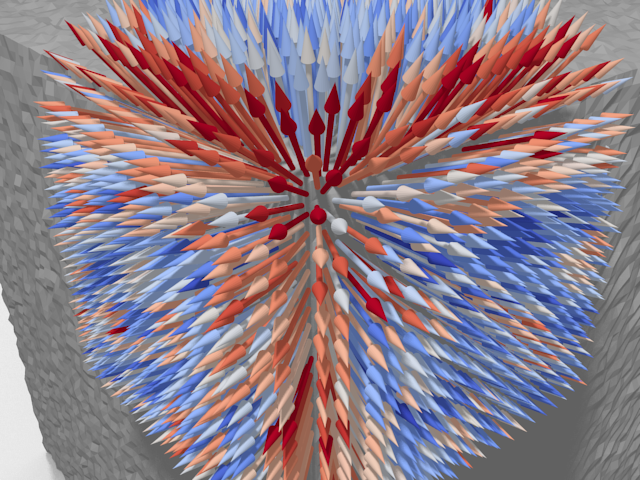}}
\subfigure[ROF denoising]{\label{fig:denoising3}\includegraphics[width=0.32\columnwidth]{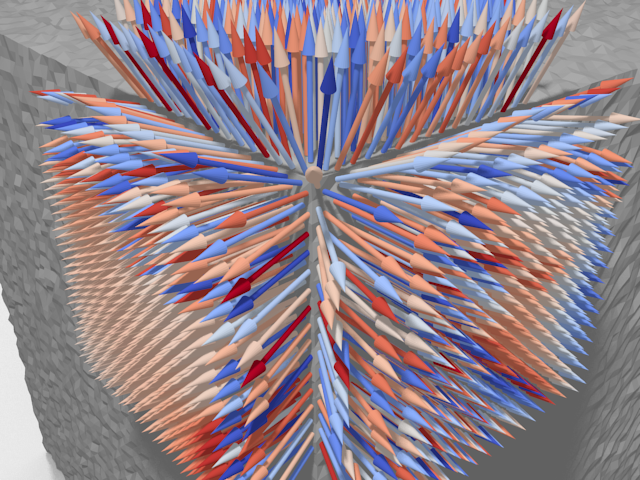}}\\
\subfigure[]{\label{fig:denoising4}\includegraphics[width=0.4\columnwidth]{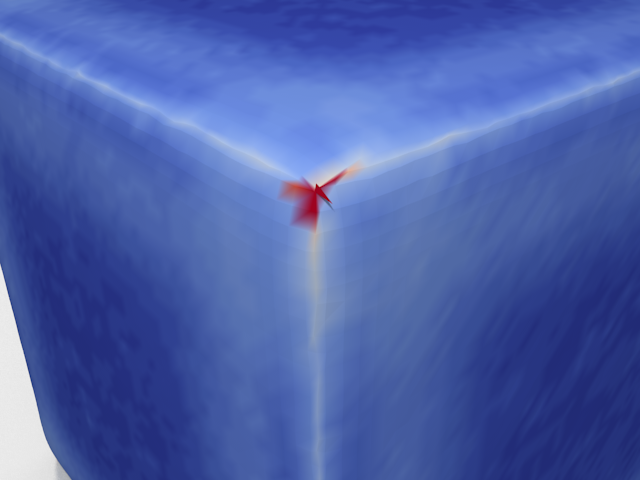}}\hspace{\myfigspacer}
\subfigure[]{\label{fig:denoising5}\includegraphics[width=0.4\columnwidth]{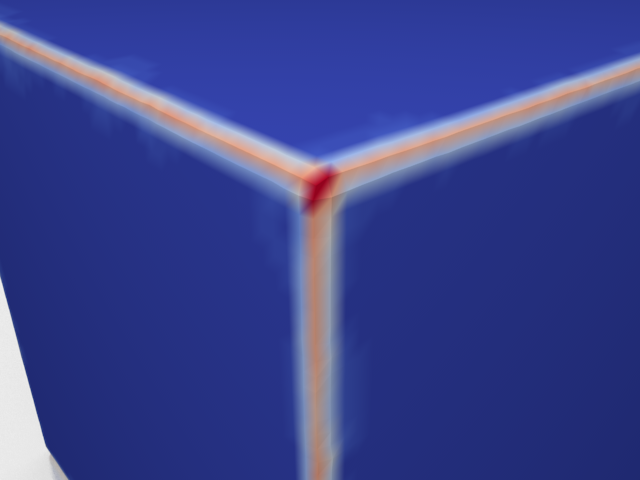}}
\caption{Integral surfaces of the normal fields in~\subref{fig:denoising2} and~\subref{fig:denoising3} obtained by LMD are shown in~\subref{fig:denoising4} respectively~\subref{fig:denoising5} together with the local residual, also see Fig.~\ref{fig:cubedenoising}.}
\label{fig:cubedenoising_close}
\end{figure}

\subsection{Normal field integration and denoising}\label{subsec:denoising}

\begin{figure}[tb]
\centering
\subfigure[Teapot]{\label{fig:convergence1}\includegraphics[width=0.4\columnwidth]{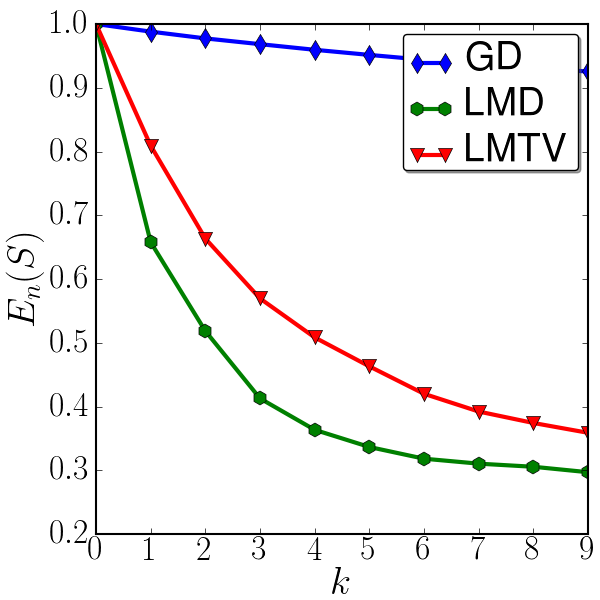}}\hspace{\myfigspacer}
\subfigure[Cube ROF]{\label{fig:convergence2}\includegraphics[width=0.4\columnwidth]{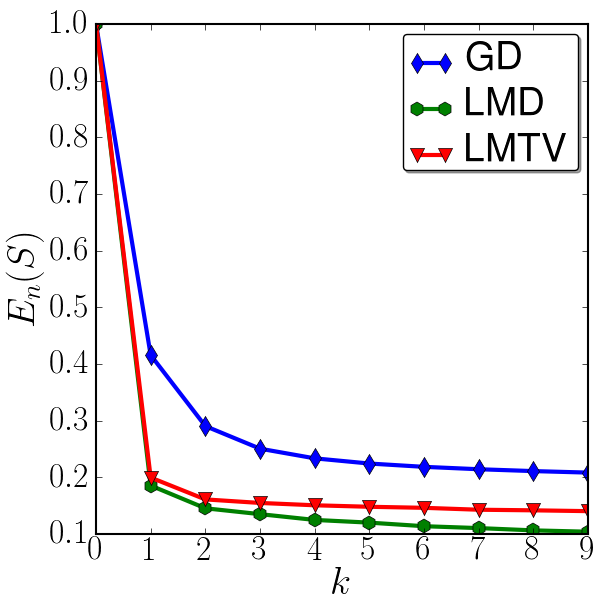}}\\
\subfigure[Sphere]{\label{fig:convergence3}\includegraphics[width=0.4\columnwidth]{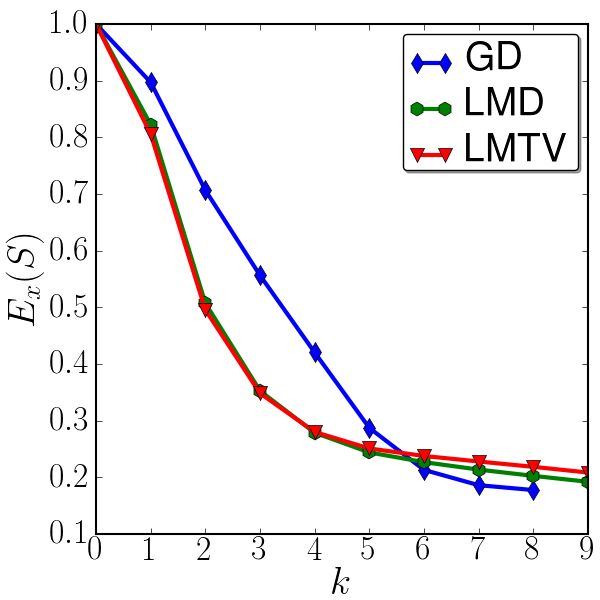}}\hspace{\myfigspacer}
\subfigure[Photometric optimization]{\label{fig:convergence4}\includegraphics[width=0.4\columnwidth]{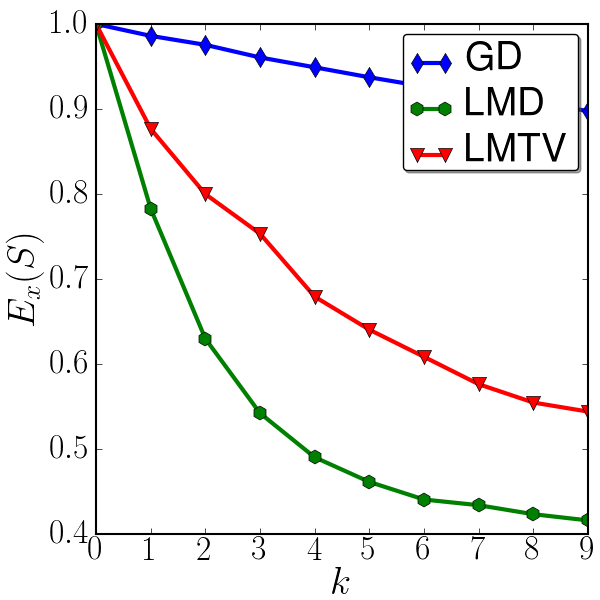}}
\caption{Convergence rates of first- vs. second-order methods.}\label{fig:convergence}
\end{figure}

Let $\bm{n}_d$ denote some \emph{desired} normal field. Integration we understand as finding a surface $S$ such that $\bm{n}(S)=\bm{n}_d(S)$. This is an inverse problem in the spirit of Sect.~\ref{subsec:motivation}: noise in the data prevents the integrability of $\bm{n}$ and hence the existence of such a strong solution. Instead, we look for a minimizer of 
\begin{equation}\label{eq:normalintegration}
	E_n(S)=\int\limits_{S}\frac{1}{2}\|\bm{n}-\bm{n}_d\|^2\,\mathrm{d}S.
\end{equation}
This energy constitutes a special case of~\eqref{eq:separability} in which $\bm{r}_n(\bm{n})=\bm{n}-\bm{n}_d$ and $\bm{r}_x(\bm{x})=\bm{0}$. It is useful in a variety of applications which are classified by how they define the target normal field $\bm{n}_d$. Take for instance the deflectometric reconstruction of specular surfaces. In deflectometry, one measures the correspondence between pixels on the image plane and points in the scene they see \emph{via} the specular surface, cf. Fig.~\ref{fig:teapot1} and~\cite{Balzer2010}. Reconstructions are shown in Fig.~\ref{fig:teapoterror}. As seen in the convergence plot in Fig.~\ref{fig:convergence1}, despite the implicit Euler integration~\eqref{eq:impliciteuler}, the gradient descent suffers from severe step size restrictions and terminates prematurely in a local minimum. Another example is fourth-order surface denoising~\cite{Avron2010}. The idea is that instead of smoothing the surface itself, which would involve a second-order diffusion equation, one first applies the smoothing to the normal field of the surface and in a second step integrates the result\footnote{In both steps, a second-order partial differential equation has to be solved, hence, we have a scheme of total order four.} $\bm{n}_d$. We obtain $\bm{n}_d$ either in terms of a maximum likelihood (ML) estimate (Fig.~\ref{fig:denoising2}) or from the output of our ADMM variant applied to the ROF-functional (Fig.~\ref{fig:denoising3}) of the original normal field (Fig.~\ref{fig:denoising1}). Remarkably, the convergence rate of GD at integration becomes competitive again given that the input data has undergone the initial smoothing (Fig.~\ref{fig:convergence2}). 

\subsection{Surface reconstruction from point clouds}\label{subsec:pclreconstruction}

\begin{figure}[tb]
\centering
\subfigure[Initialization]{\label{fig:sphere_init}\includegraphics[width=0.24\columnwidth]{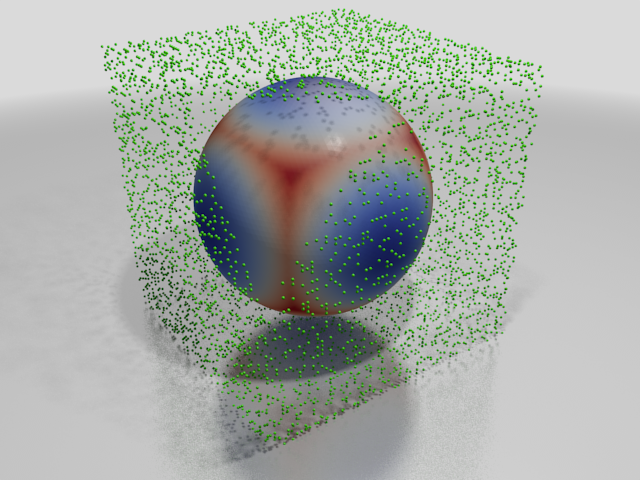}}
\subfigure[GD]{\label{fig:sphere_grad}\includegraphics[width=0.24\columnwidth]{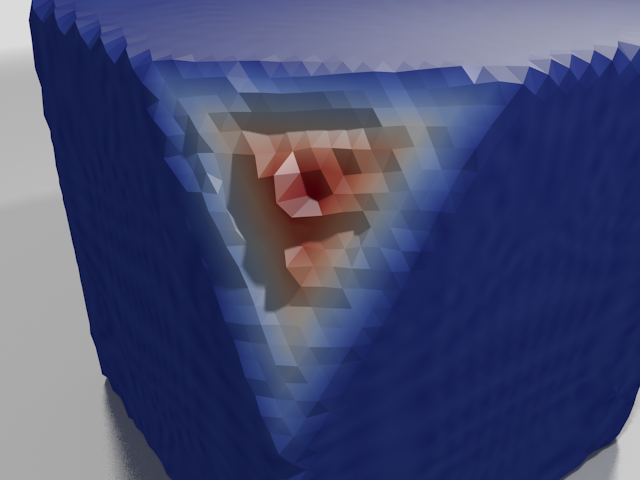}}
\subfigure[LMD]{\label{fig:sphere_lm}\includegraphics[width=0.24\columnwidth]{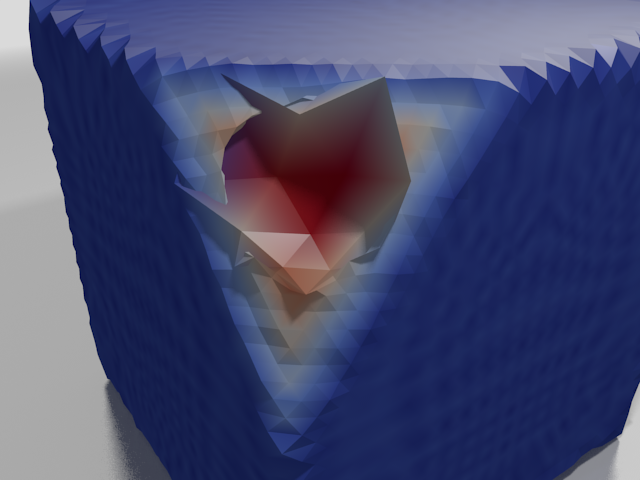}}
\subfigure[LMTV]{\label{fig:sphere_sb}\includegraphics[width=0.24\columnwidth]{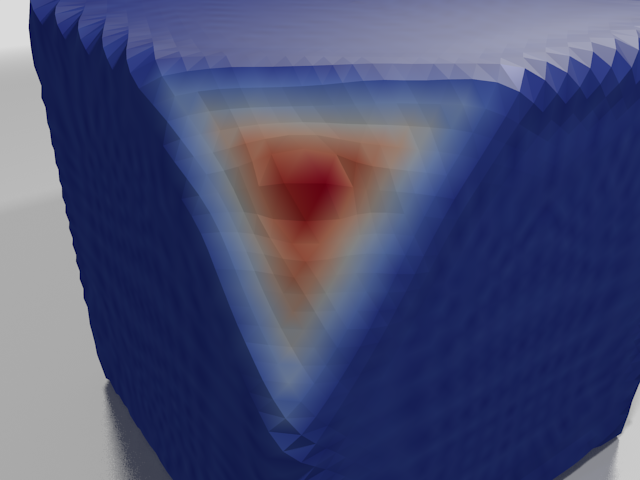}}
\caption{\subref{fig:sphere_init} To interpolate a box-shaped point cloud with one corner cut off, we evolve a sphere towards minimal average distance. \subref{fig:sphere_grad} GD suffers from a catastrophic loss of stability after $10$ steps, \subref{fig:sphere_lm} LMD develops folds after $30$, while \subref{fig:sphere_sb} LMTV remains stable.}\label{fig:spherestability}
\end{figure}

\begin{figure}[tb]
\centering
\subfigure[Point cloud]{\label{fig:kinect1}\includegraphics[width=0.43\columnwidth]{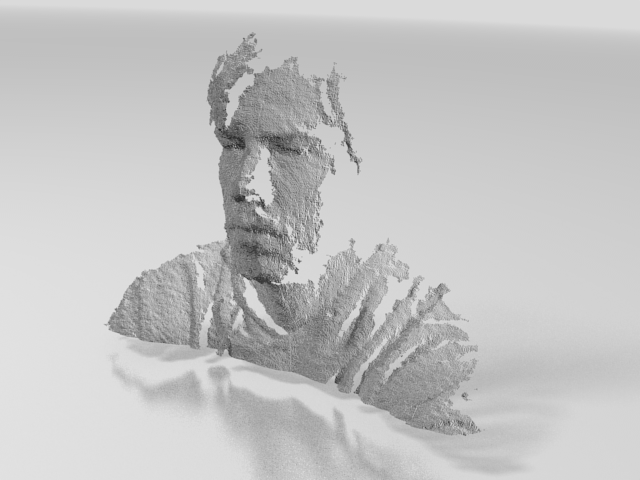}}\hspace{0.2cm}
\subfigure[Normal map]{\label{fig:kinect2}\includegraphics[width=0.43\columnwidth]{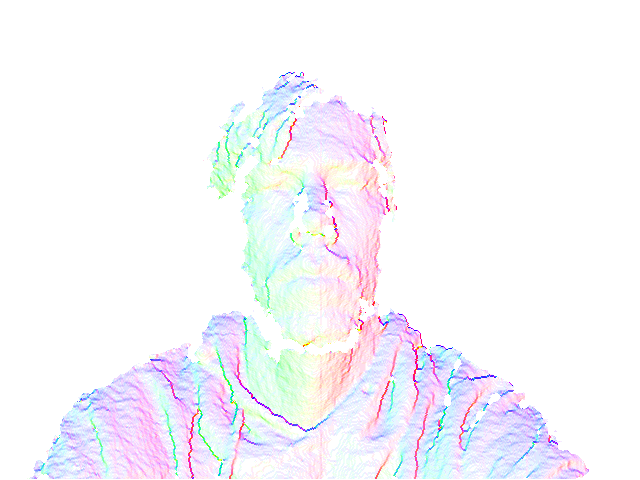}}\\
\subfigure[Initialization]{\label{fig:kinect3}\includegraphics[width=0.43\columnwidth]{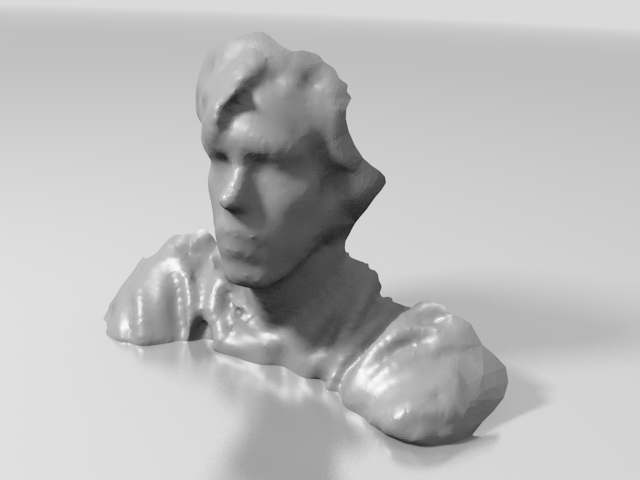}}\hspace{0.2cm}
\subfigure[LMTV refinement]{\label{fig:kinect4}\includegraphics[width=0.43\columnwidth]{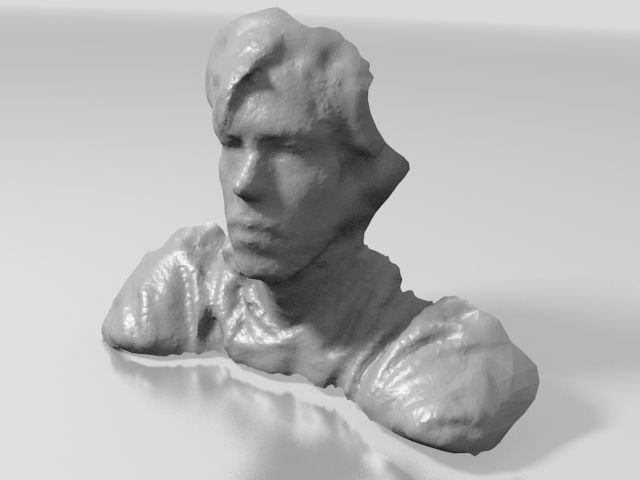}}
\caption{Hermite interpolation of points from an RGBD camera.}
\label{fig:kinect}
\end{figure}

Here, we are given a set of discrete points in space $\mathcal{P}=\{\bm{p}_l\in\mathbb{R}^3\;|\; l\in\mathbb{N}\}$ (Fig.~\ref{fig:kinect1}), which are the representation of choice for many reconstruction methodologies embracing the triangulation principle. Our goal is to find a surface $S$ with minimal average distance
\begin{equation}\label{eq:pointdistance}
	E_x(S)=\int\limits_S \frac{1}{2}\|\bm{x}-\hat{\bm{x}}\|^2\,\mathrm{d}\bm{x}
\end{equation}
to $\mathcal{P}$ where $\hat{\bm{x}}=\arg\min_{\bm{p}_l\in\mathcal{P}}\|\bm{x}-\bm{p}_l\|$. Making the substitution $\bm{r}_x(\bm{x})=\bm{x}-\hat{\bm{x}}$ respectively $\cost(\bm{x})=\frac{1}{2}\|\bm{r}_x\|^2$, this energy can be brought into the form~\eqref{eq:separability} with $\bm{r}_n$ vanishing. Supposing that $\mathcal{P}$ is sufficiently dense, the global minimum with value $0$ is given by the zero-set $\cost^{-1}(0)$ of $\cost(\bm{x})$. This direct approach requires a representation of the squared distance function over a Cartesian grid. Regularity and dimensionality of such a representation imply a tradeoff between reconstruction quality and computational efficiency\footnote{Insufficent spatial resolution is known to cause so-called \emph{staircasing artifacts} at the numerical extraction of the level set.}, making it difficult to take advantage of the full resolution of the raw data. While it lacks fine geometric details, $\cost^{-1}(0)$ generally captures the topology of the surface we wish to infer, thus providing an adequate initial guess $S_0$ for refinement by~\eqref{eq:shapegaussnewton}. The Jacobian $D\bm{r}_x|_{\bm{x}}$ is given by the vector that connects $\bm{x}$ with its closest point $\hat{\bm{x}}\in\mathcal{P}$. If the point cloud is oriented such that for each $\bm{p}_l\in\mathcal{P}$, we have a desired orientation $\bm{n}_d$, we can combine~\eqref{eq:pointdistance} and~\eqref{eq:normalintegration} with $\bm{r}_n(\bm{x})=\bm{n}(\bm{x})-\bm{n}_d(\hat{\bm{x}})$ similar to a \emph{screened Poisson reconstruction}~\cite{Kazhdan2013}. To test the performance of the different algorithms under ideal circumstances, we synthesized the toy example shown in Fig.~\ref{fig:sphere_init}. We observe at hand of Fig.~\ref{fig:spherestability} that LMTV is the only method that achieves a stable stationary state, which justifies the TV as a regularizer. Again, the error decay in the GD method is satisfactory, which can be attributed to the ideal circumstances and that the Euler steps are backwards. A more realistic scenario is depicted in Fig.~\ref{fig:kinect}. Here, the point cloud stems from an RGBD sensor. Given the known and regular topology of the image lattice, one can easily obtain a normal map for each depth image by finite-differencing (Fig.~\ref{fig:kinect2}). We obtain and initial reconstruction by Poisson reconstruction  (Fig.~\ref{fig:kinect3}) and refine the level of detail by minimizing a blend of the functionals in~\eqref{eq:normalintegration} and~\eqref{eq:pointdistance}. Thereby, the tradeoff between point and normal fidelity can be steered by a scalar weight. The outcome is shown in Fig.~\ref{fig:kinect4}.  

\begin{figure}
\centering
\subfigure[$k=0$]{\label{fig:mvslm00}\includegraphics[width=0.45\columnwidth]{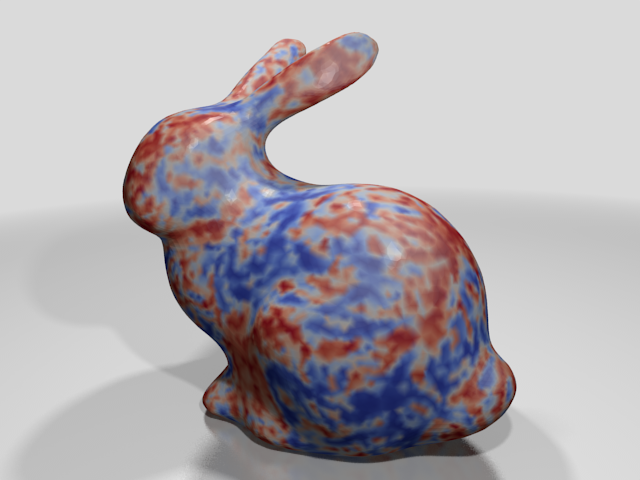}}\hspace{\myfigspacer}
\subfigure[$k=1$]{\label{fig:mvslm01}\includegraphics[width=0.45\columnwidth]{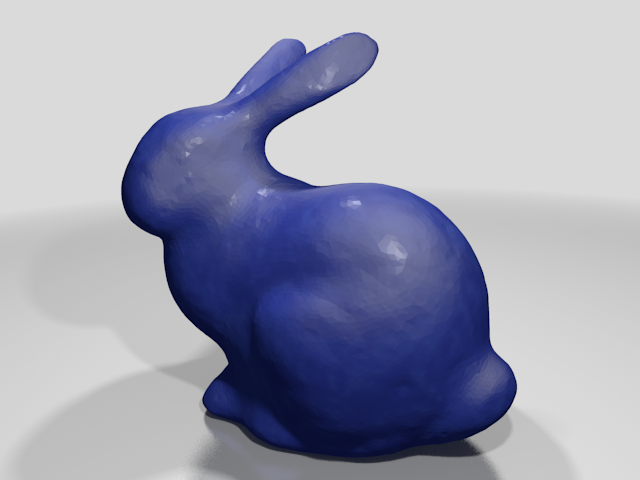}}
\caption{The second-order methods take large steps towards the minimal re-projection error, whose local value is shown color-coded (normalized w.r.t. its peak value at $k=0$).}\label{fig:mvslm}
\end{figure}

\subsection{Photometric optimization}\label{subsec:mvs}

\begin{figure}[tb]
\centering
\subfigure[$k=0$]{\label{fig:shrinking1}\includegraphics[width=0.32\columnwidth]{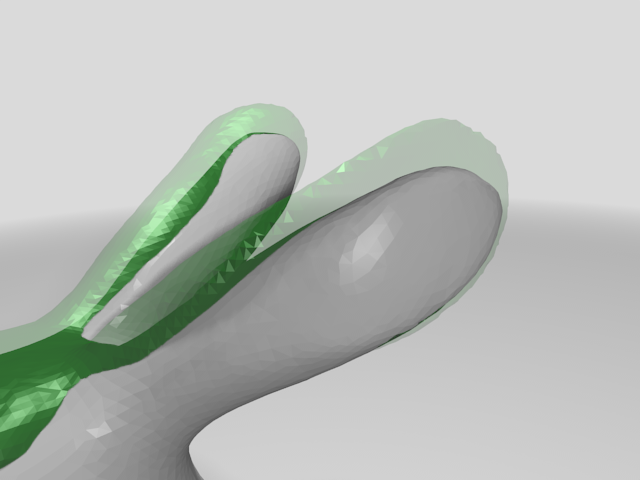}}\hfill
\subfigure[$k=5$]{\label{fig:shrinking2}\includegraphics[width=0.32\columnwidth]{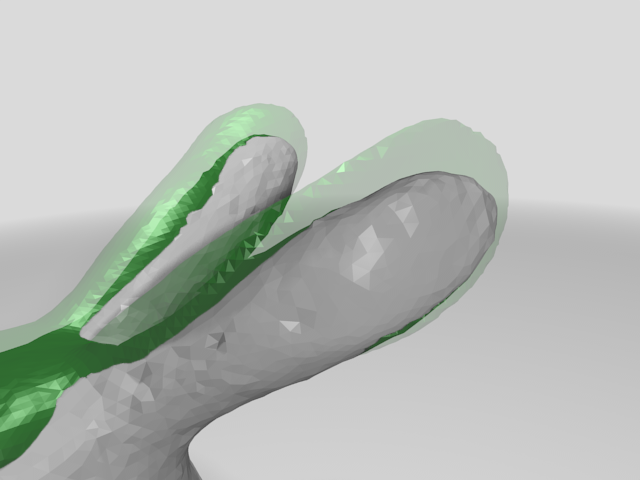}}\hfill
\subfigure[$k=48$]{\label{fig:shrinking3}\includegraphics[width=0.32\columnwidth]{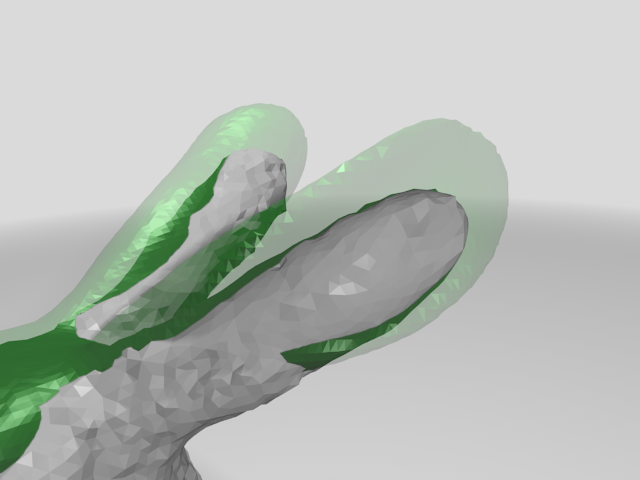}}
\caption{Plotting the surface evolved by gradient descent relative to the ground truth model (green) reveals the well-known and undesired effects of MCM.}\label{fig:shrinking}
\end{figure}

\begin{figure}[tb]
\centering
\subfigure[Initialization]{\label{fig:mvsresult1}\includegraphics[width=0.45\columnwidth]{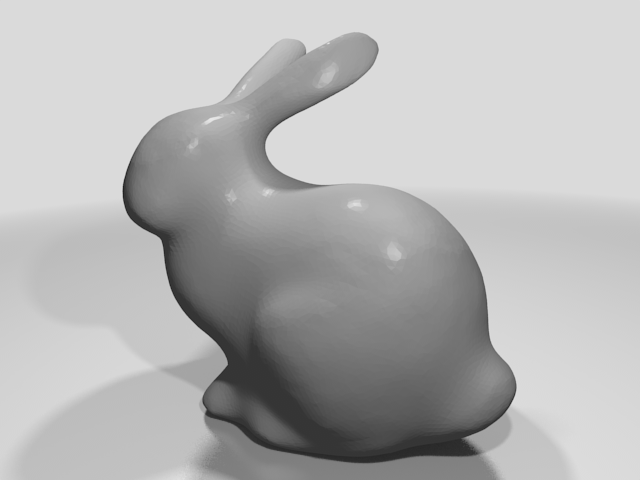}}\hspace{\myfigspacer}
\subfigure[GD after $48$ steps]{\label{fig:mvsresult2}\includegraphics[width=0.45\columnwidth]{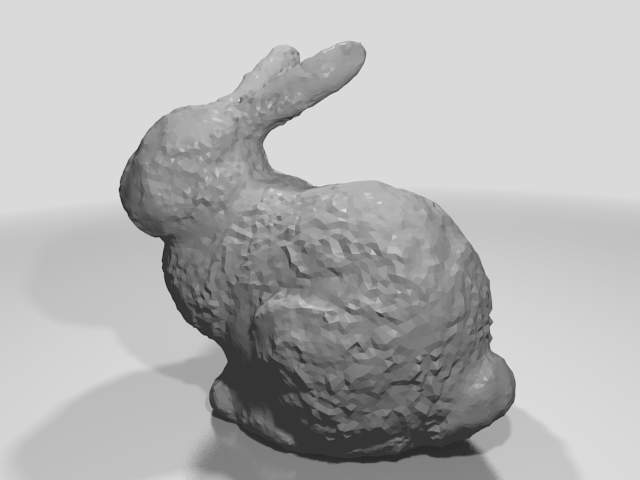}}\\
\subfigure[LMD at $k=5$]{\label{fig:mvsresult3}\includegraphics[width=0.45\columnwidth]{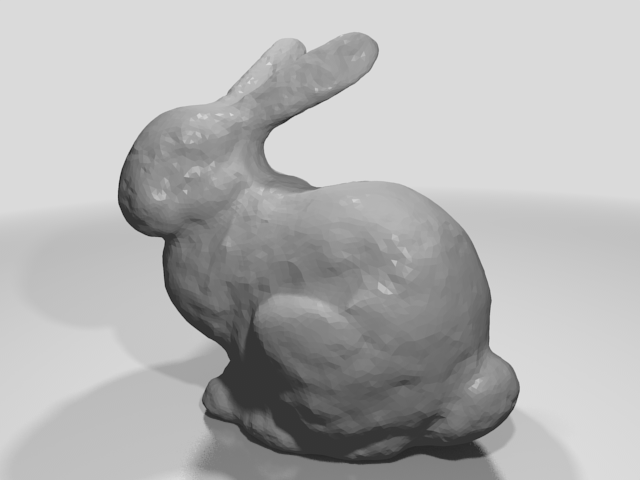}}\hspace{\myfigspacer}
\subfigure[LMTV at $k=5$]{\label{fig:mvsresult4}\includegraphics[width=0.45\columnwidth]{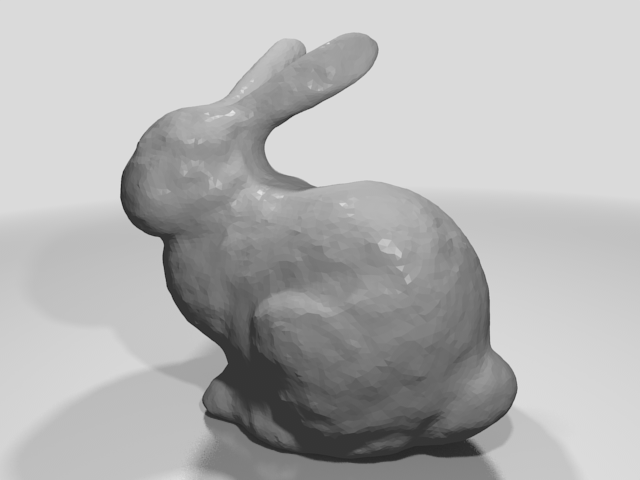}}
\caption{Photometric optimization results.}\label{fig:mvsresult}
\end{figure}

Suppose we have a Lambertian surface of which we capture a set of gray value images from different vantage points. Multiview stereo is concerned with the inverse problem of converting the data into a geometric model of the surface. We cannot delve into the details of this highly sophisticated process. We limit the discussion to another application of our algorithm in a stage at the very end of the reconstruction pipeline, i.e., after an initial approximation of the surface as a set of (oriented) points has been armed with the topological structure of a surface. \emph{Photometric optimization} seeks a minimizer of the shape functional 
\[
	E_x(S)=\int\limits_S \frac{1}{2}\rho^2\,\mathrm{d}\bm{x}
\]
where $\rho$ measures the instantaneous photoconsistency between pairs of images. Generally, $\rho$ depends on shape and radiometry of the unknown surface as well as the set of vantage points. Its precise form used in our experiments is described in Appx.~\ref{appx:mvs}. Let us only remark that it does not incorporate additional knowledge on the location of the contour generators.  We call $\rho$ \emph{instantaneous} because $\rho$ also depends on \emph{visibility}, which can only be modelled numerically, but not analytically. In local shape optimization, this typically happens at each iteration. Due to aforementioned complexity of the problem, we chose to study our algorithm in a controllable test scenario. We rendered synthetic images of the Stanford bunny model outfitted with a random texture (Fig.~\ref{fig:mvsintro}). As it is standard, we estimated the visual hull for an initialization but applied aggressive Laplacian smoothing to it to obtain a surface further away from the minimizer. The deviation between initial and ground truth model can be seen in Fig.~\ref{fig:shrinking1}. The initial surface along with our reconstruction results is shown in Fig.~\ref{fig:mvsresult}. Needless to say, even under these conditions, perfect recovery of the ground truth model is all but impossible as it crucially depends on sufficient texture and sampling. Figs.~\ref{fig:mvslm} and~\ref{fig:convergence4} confirm the superior convergence rate of LMTV and LMD. The convergence behavior of GD is similarly bad as in the teapot example, although here, the input data should be far less challenging. The stiffness of the evolution equation~\eqref{eq:impliciteuler} and thus the maximal attainable step size is determined by the value of the regularization weight $\lambda$. In all our experiments, we were forced to set $\lambda$ to extremely high values for GD to maintain stability yet were obtaining unusually rugged reconstructions such as in Fig.~\ref{fig:mvsresult2}. Finally, let us remark that evaluating $\rho$ generates the majority of computational cost at each time step (Tab.~\ref{tab:steptimes}). Since, independent of the optimization order, this cost scales quadratically in the number of views, superlinear convergence becomes critically important in the present application. 

\section{Conclusion and future work}

We have presented a general second-order optimization method for shape functionals with several applications in the realm of visual reconstruction. Apart from numerical feasibility studies, our contribution is more theoretical. We hope to pave the way for second-order methods in shape optimization but wish to further investigate their practical relevance in future work ourselves.

\newpage

\appendix

\section{LS shape functionals depending on curvature}\label{appx:curvature}

The ideas developed in Sect.~\ref{subsubsec:shapelm} of the main paper extend to separable and quadratic functionals of the curvature. There are many different notions of curvature. We first show that the mean curvature is particularly representative. For this, we need to recall a few definitions, which can be found in~\cite{DoCarmo1976} among several other sources: The surface Jacobian of the normal field $D_{S}\bm{n}$ is equivalent to the \emph{shape operator} or \emph{Weingarten map} on $S$ (up to an additional minus sign). It serves as the Gram matrix of the \emph{second fundamental form} $\mathrm{II}:(\bm{s},\bm{t})\mapsto \langle -D_{S}\bm{n}\bm{s},\bm{t}\rangle$, which can be used to measure the sectional curvature of $S$ in the tangential direction $\bm{t}$ via $\mathrm{II}(\bm{t},\bm{t})$. In view of the connection between $\mathrm{II}$ and $D_S\bm{n}$, 
\[
	E(S)=\int\limits_{S}\frac{1}{2}\|D_{S}\bm{n}\|^2_{\mathrm{F}}\, \mathrm{d}S
\]
fully describes the \emph{bending energy} of $S$. One can simplify $E(S)$ further by removing its dependence on the Gaussian curvature $\gamma$: From the identity $\|D_{\Gamma}\bm{n}\|^2_{\mathrm{F}}=\kappa^2-2\gamma$, it immediately follows
\[
	E(S)=\int\limits_{S}\frac{1}{2}\kappa^2\,\mathrm{d}S-\int\limits_{S}\gamma \,\mathrm{d}S=\int\limits_{S}\frac{1}{2}\kappa^2\,\mathrm{d}S-\pi\chi(S).
\] 
The last equality is a direct consequence of the \emph{Gauss-Bonnet theorem}, which states that the total Gaussian curvature of a compact regular surface equals $2\pi$ times its \emph{Euler characteristic} $\chi(S)$.  During the quest for a stationary state, one might as well drop this constant term. The remaining summand is the so-called \emph{Willmore functional}
\begin{equation}\label{eq:willmore}
    E_W(S)=\int\limits_{S}\frac{1}{2}\kappa^2\,\mathrm{d}S.
\end{equation}
This reduction nurtures the hope that it may be possible to express all shape functionals depending on curvature in terms of $\kappa$, for which the shape differential $\partial_v\kappa =-\Delta_S v$ is known~\cite{Delfour2001}. If we denote by $\bm{r}_c$ the residual that only depends on $\kappa$ but not on $\bm{x}$ and $\bm{n}$,  we get the following extension of Eq.~(9) in the main paper:
\begin{multline*}\label{eq:localshapefunction}
	E_d(v):=\frac{1}{2}\|\bm{r}_x(\bm{x})+D\bm{r}_x(v\bm{n})\|^2_{L^2(S)}\\+\frac{1}{2}\|\bm{r}_n(\bm{n})-D_{\mathbb{S}^2}\bm{r}_n\nabla_S v\|^2_{L^2(S)}
	+\frac{1}{2}\|\bm{r}_{c}(\kappa)-\partial_{\kappa}\bm{r}_{c}\Delta_S v\|^2_{L^2(S)}.
\end{multline*}
In the example of the Willmore energy~\eqref{eq:willmore}, we have $\bm{r}_{c}(\kappa)=\kappa$ and $\partial_{\kappa}\bm{r}_{c}(\kappa)=1$. In comparison, the first-order shape differential of $E_W$ is known to have the following form:
\[
	DE_W(S;v)=\int\limits_S\kappa\left(\langle\nabla\kappa,\bm{n}\rangle+\frac{1}{2}\kappa^2\right)v-\kappa \Delta_S v\,\mathrm{d}S,
\]
cf.~\cite{Droske2004}. Curvature-dependent functionals may play a role e.g. in surface fairing~\cite{Bobenko2005} or the inference of specular flow\footnote{Like the law of reflection itself, specular flow is mainly influenced by the Gauss map of the mirror respectively its spatial and temporal changes, the former being closely related to the curvature of the surface.}~\cite{Adato2010}.

\section{Finite-elements analysis on triangular meshes}\label{appx:fem}

\begin{figure}
\subfigure[Face]{\label{fig:triangle}\def\svgwidth{0.3\columnwidth}
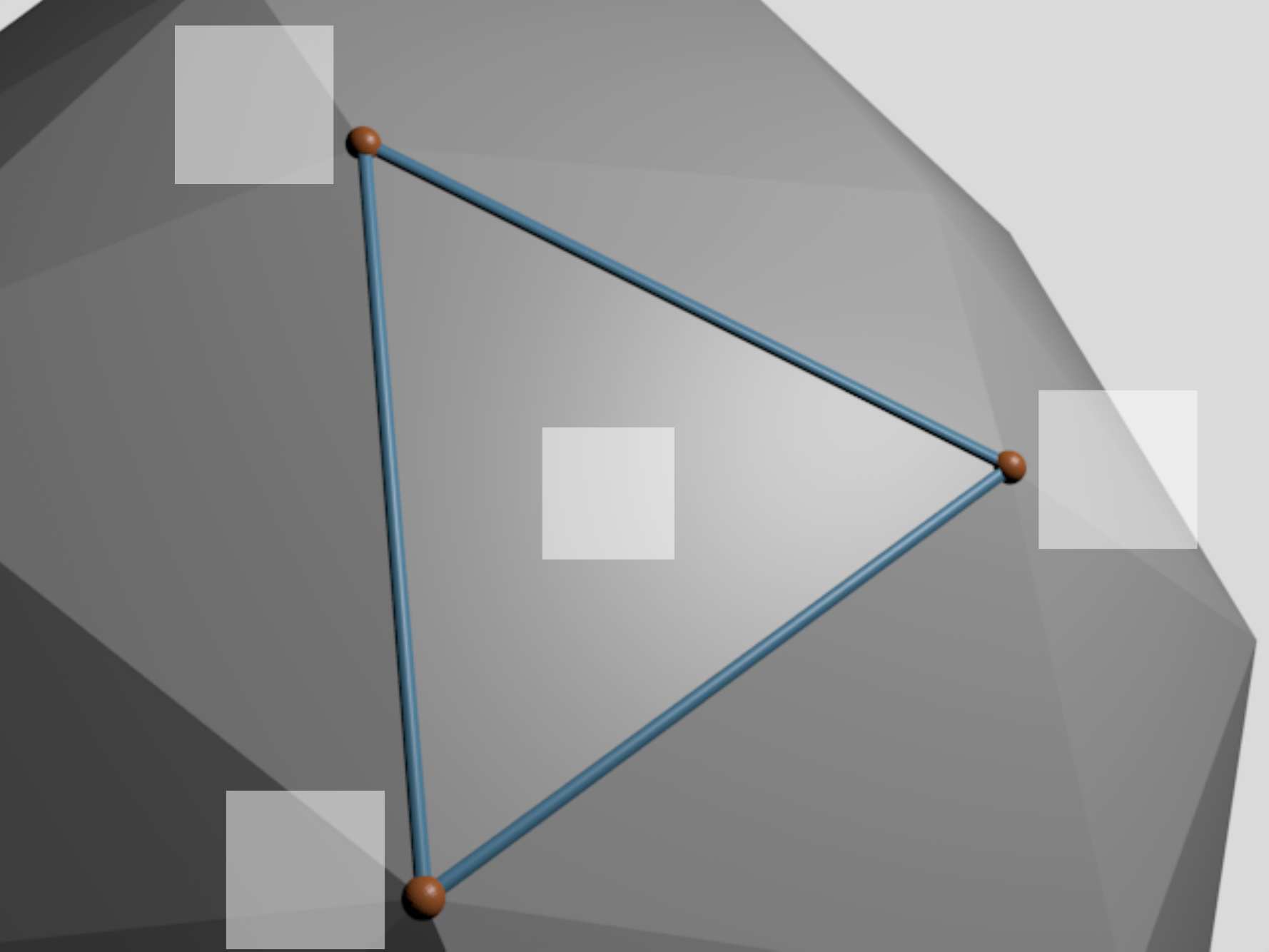}
\subfigure[Dual edge vector]{\label{fig:dualedge}\def\svgwidth{0.3\columnwidth}
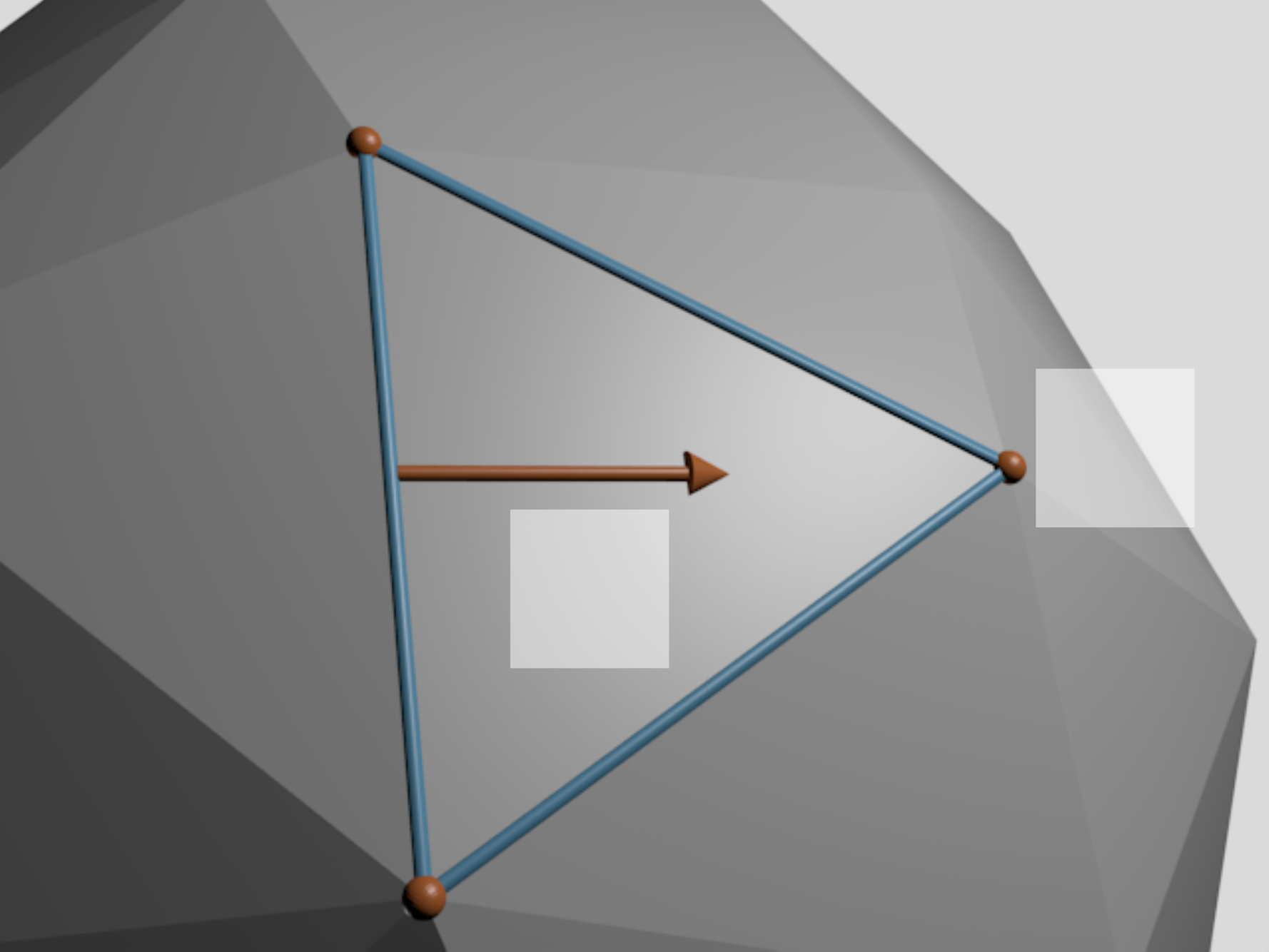}
\subfigure[One-ring neighborhood]{\label{fig:onering}\def\svgwidth{0.3\columnwidth}
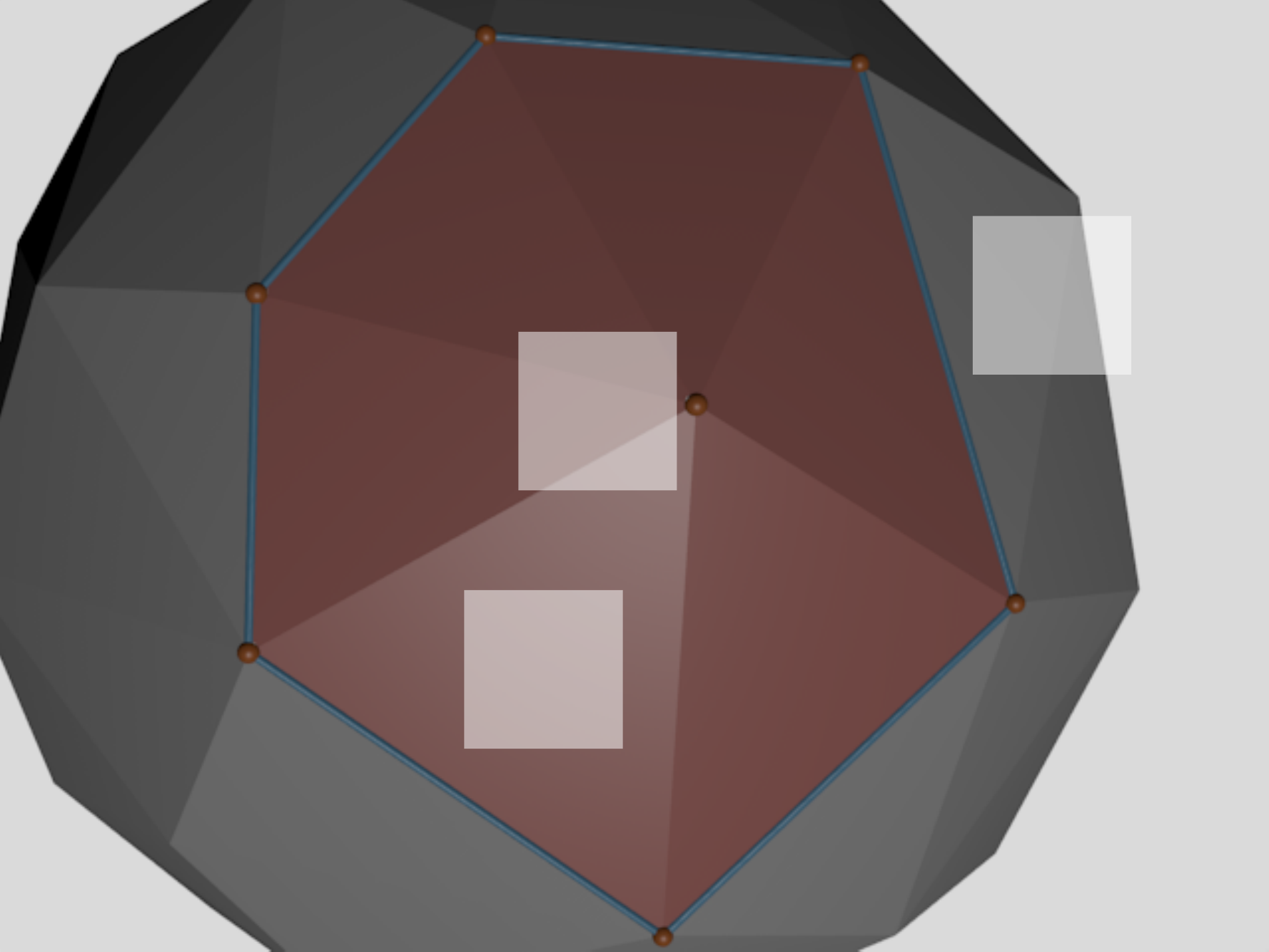}
\caption{Local mesh geometry.}\label{fig:meshgeometry}
\end{figure}

In our effort to support reproducible research, we provide all necessary details to implement the ADMM on triangle meshes described in Sect.~\ref{subsec:sbcgls} of the main paper. We model the tangent space $H^1(S)$ respectively $BV(S)$ of $X=\diff(S_0,\mathbb{R}^3)/\diff(S_0,S_0)$ by the space of conforming finite elements
\[
	D_h=\{v_h:S_h\to\mathbb{R}\;|\;v_h\in C(S_h),\; v_h\mbox{ linear on all } T\},			
\]
where by $T$, we denote the faces of the mesh. Clearly, $(D_h,\|\cdot\|_{D_h})$ is an $n$-dimensional Hilbert space spanned by the basis of piecewise linear hat functions $\varphi_i$ taking the value $1$ on each vertex $\bm{x}_i\in S_h$ and vanishing on the boundary edges of its one-ring neighborhood $\mathcal{N}(\bm{x}_i)$, in short $\varphi_i(\bm{x}_j)=\delta_{ij}$. Every element of $D_h$ can be written as the unique linear combination
\[
	v_h=\sum_{i=1}^n v_i\varphi_i, \quad v_i=v_h(\bm{x}_i)\in\mathbb{R}\quad\forall i\in\{1,\ldots,n\}.
\]
Its gradient exists on each triangle and is constant there:
\[
	\nabla_{S_h} v_h|_T =\frac{1}{2|T|}\sum_{i_k\in\mathcal{I}(T)} v_{i_k}\bm{e}_{i_k}^{\perp}.
\]
Here, $|T|$ is the area of $T$, the set $\mathcal{I}(T)$ indexes the vertices which form the triangle, and $\bm{e}_{i_k}$ is the vector orthogonal to both, $\bm{n}(T)$ and the edge opposite to the vertex $\bm{x}_{i_k}$ within $T$, cf.~\cite{Pinkall1993} and Fig.~\ref{fig:triangle}-\subref{fig:dualedge}. This relation defines a discrete gradient operator on the entire mesh which -- with slight overloading of notation -- we write as $\nabla_{S_h}\in\mathbb{R}^{m\times n}$, where $m$ is three times the number of faces.

We need to be able to express $L^2(S_h)$-norms of a function $v_h$ in terms of a weighted $\ell^2$-norm of the vector $\mathbf{v}\in D_h$. To this end, we perform \emph{Gauss-Legendre quadrature} over each triangle:
\[
	\|v_h\|^2_{L^2(S_h)}=\sum_{T\in S_h}\frac{|T|}{3} \sum_{i_k\in\mathcal{I}(T)} v_{i_k}^2.
\]
It follows that $\|v_h\|^2_{L^2(S_h)}=\|\mathbf{W}_x\mathbf{v}\|^2_{\ell^2}=\|\mathbf{v}\|^2_{D_h}$ . The matrix $\mathbf{W}_x\in\mathbb{R}^{n\times n}$ weighing the $\ell^2$-norm (called \emph{lumped mass matrix} in the finite-elements literature) contains the sum of the areas of all triangles in $\mathcal{N}(\bm{x}_i)$:
\[
	\mathbf{W}_x=\diag\left(\sqrt{w_{x,i}}\right),\quad w_{x,i}=\frac{1}{3}\sum\limits_{T_j\in\mathcal{N}(\bm{x}_i)}|T_j|, 
\]
see Fig.~\ref{fig:onering}. Similarly, 
\[	
	\|\nabla_{S_h} v_h\|^2_{L^2(S_h)}= \sum_{T\in S_h} |T|\|\nabla_{S_h} v_h|_T\|^2
\]
implies $\mathbf{W}_n=\diag(\sqrt{|T|})$ for the face-based mass matrix $\mathbf{W}_n\in\mathbb{R}^{m\times m}$.

\begin{figure}
\centering
\subfigure[Hernandez's function for different degrees $\sigma$ of smoothing.]{\label{fig:hernandez}\includegraphics[width=0.4\textwidth]{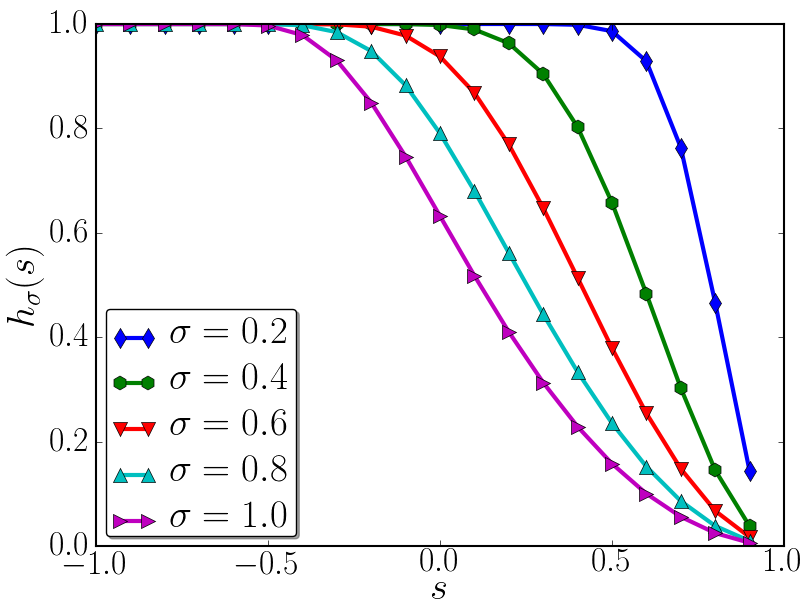}}\hspace{0.5cm}
\subfigure[Dino sparse ring data set]{\label{fig:dinodata}\includegraphics[width=0.5\textwidth]{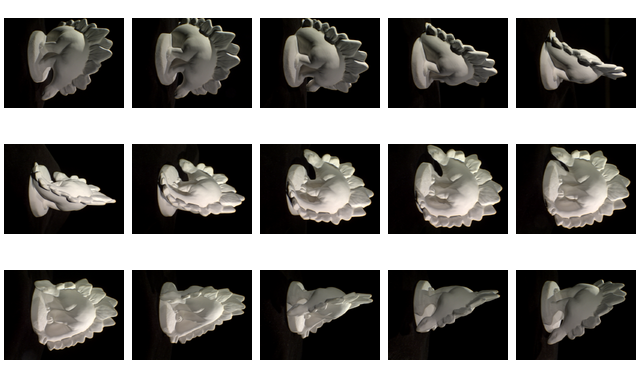}}
\caption{Multiview stereo and photometric optimization.}\label{fig:mvsappendix}
\end{figure}

\section{Photometric optimization}\label{appx:mvs}

\subsection{Derivation of the re-projection error}

The residual in this example arises from the following chaining 
\[
	r_x(\bm{x};s,t)=h_{\sigma}\circ\ncc(\bm{x};s,t)
\]
of a function  $h_{\sigma}:[-1,1]\to [0,1]$ proposed in~\cite{Hernandez2004} and some normalized cross-correlation (NCC). Besides the scene point $\bm{x}$, the residual depends on the indices $s,t\in\mathbb{N}$ of two views in which $\bm{x}$ is visible. Hernandez's function
\[
	h_{\sigma}(s) := 1-\exp\left(-\frac{\left[\tan\left(\frac{\pi}{4}(s-1)\right)\right]^2}{\sigma^2}\right)
\]
compresses the range of the correlation coefficient to the interval $[0,1]$, and since  $\partial_s h_{\sigma}<0$, does so in an orientation-reversing fashion. Meanwhile, it attenuates or suppresses high errors depending on the choice of $\sigma$, see Fig.~\ref{fig:hernandez}. The normalized cross-correlation $\ncc:\mathbb{R}^d\times\mathbb{R}^d\to [-1,1]$, 
\[
	\ncc(\bs{\varphi}_s,\bs{\varphi}_t):=\frac{\langle\bs{\varphi}_s,\bs{\varphi}_t\rangle}{\|\bs{\varphi}_s\|\|\bs{\varphi}_t\|},
\]
is computed over the values $\bs{\varphi}_s,\bs{\varphi}_t$ of two local image descriptors $\varphi_s,\varphi_t$ such as e.g. Histograms of Oriented Gradients (HOG) or local patches of the images themselves. More precisely, if we model vantage points by elements $g$ in the Euclidean group $\SE(3)$ and denote the canonical pinhole projection by $\pi$, then $\varphi_s$ maps the image $\mathcal{I}_s$ in an $\epsilon$-environment $U$ of $\pi\circ g_s(\bm{x})$ to some $d$-dimensional feature space: $\varphi_s:\mathcal{I}_s|_{U}\to\mathbb{R}^d$. In our implementation, we use the OpenGL $z$-buffer to re-project this regularly-shaped neighborhood $U$ into the image $t$ along the current surface $S$, i.e.,  $\varphi_t:\mathcal{I}_t|_{\tilde{U}}\to\mathbb{R}^d$ where $\tilde{U}=\pi\circ g_ t \circ g_s^{-1}\circ \pi_s^{-1}(U; S)$. Note that the so-obtained correlation coefficient is not symmetric w.r.t. $s$ and $t$, but this can be easily remedied by concatenating two residual vectors, one for each permutation of $(s,t)$. The same can be done to facilitate more than a single image pair (under appropriate normalization). The practical value of $\epsilon$ depends on how far the initial shape is from the desired one because we must guarantee that $U$ on $\tilde{U}$ contain the projections of a minimum number of points that were \emph{co-visible at the time of data acquisition}. In our experiments, we chose $\epsilon$ in the range of $[3,10]$.

\begin{figure}[tbp]
\centering
\subfigure[$k=0$]{\label{fig:coarseinit}\includegraphics[width=0.45\textwidth]{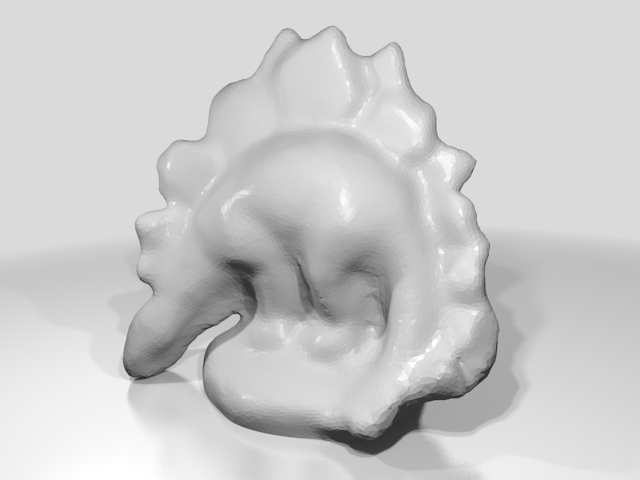}}\hspace{0.15cm}
\subfigure[$k=1$]{\label{fig:coarse01}\includegraphics[width=0.45\textwidth]{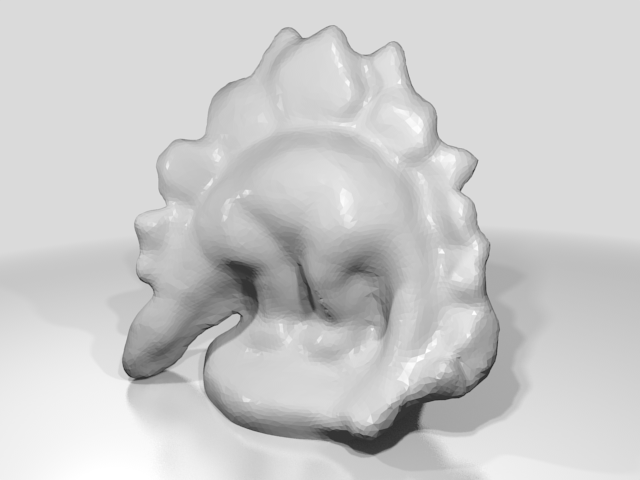}}\\
\subfigure[$k=2$]{\includegraphics[width=0.45\textwidth]{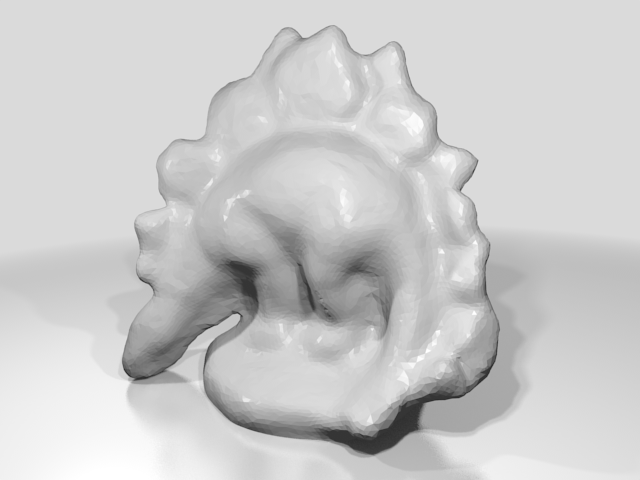}}\hspace{0.15cm}
\subfigure[$k=5$]{\label{fig:coarse05}\includegraphics[width=0.45\textwidth]{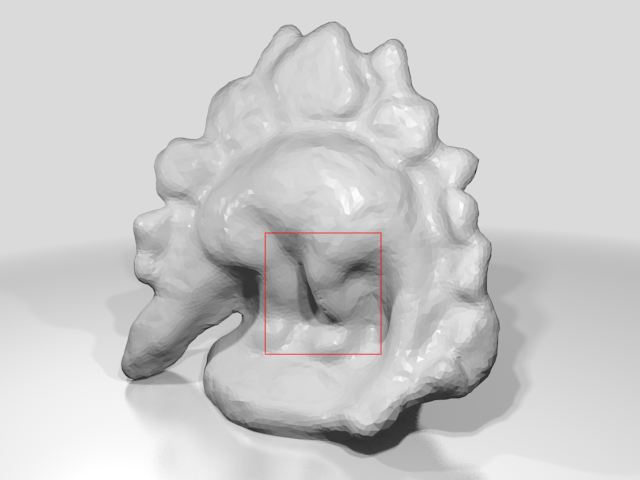}}
\caption{An initial mesh~\subref{fig:coarseinit} can be obtained from the Poisson reconstruction of an oriented point cloud~\cite{Kazhdan2013}, which we estimate by the patch-based triangulation technique proposed in~\cite{Furukawa2010}. Figs.~\subref{fig:coarse01}-\subref{fig:coarse05} illustrate the effect of $5$ LMD steps. The size of the shadows in the red-marked region in~\subref{fig:coarse05} reveals that noticeable changes of the initial shape take place. Additional supplemental material contains an animated version of this and other sequences.}\label{fig:coarseiter}
\end{figure} 

\subsection{Additional experimental results}

\begin{figure}[tbp]
\centering
\subfigure[$k=0$]{\includegraphics[width=0.19\textwidth]{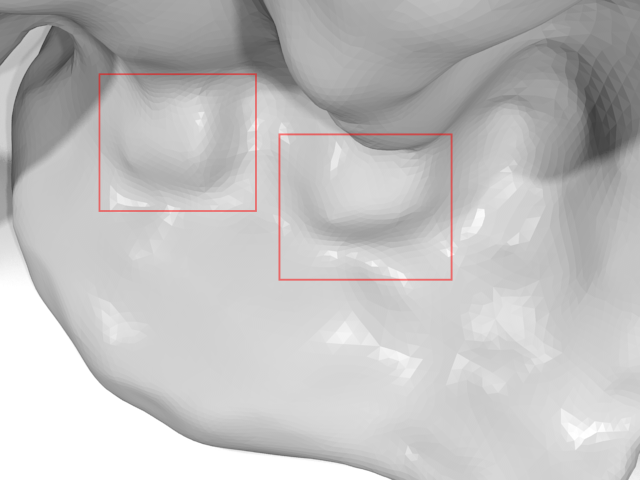}}
\subfigure[$k=1$]{\includegraphics[width=0.19\textwidth]{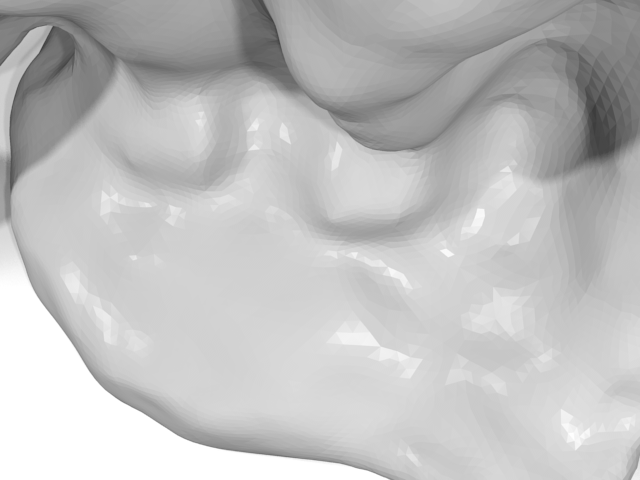}}
\subfigure[$k=2$]{\includegraphics[width=0.19\textwidth]{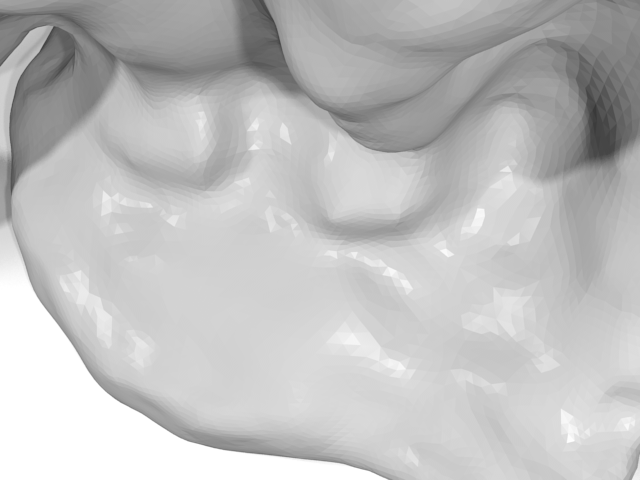}}
\subfigure[$k=3$]{\includegraphics[width=0.19\textwidth]{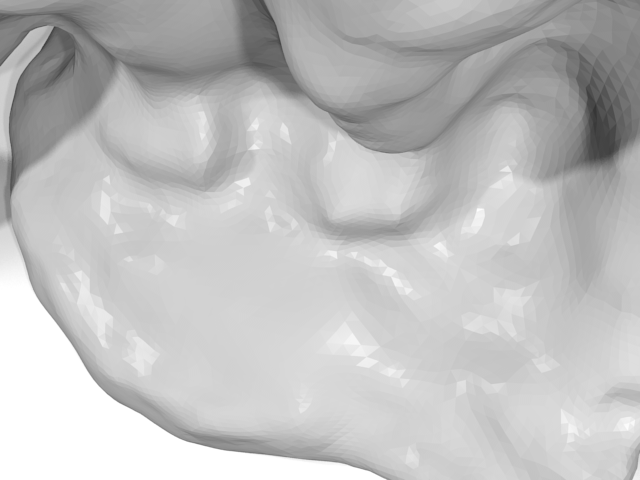}}
\subfigure[$k=5$]{\includegraphics[width=0.19\textwidth]{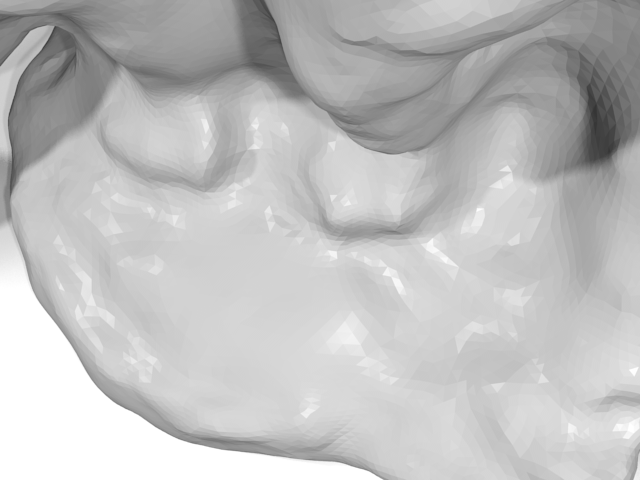}}
\caption{Additional LMD iterations of the refined model enhance small details like the dino's toes.}\label{fig:feet}
\end{figure}

We also ran our algorithm on the dino sparse ring data set described in~\cite{Seitz2006}, also see Fig.~\ref{fig:dinodata}. The results are shown and discussed in Figs.~\ref{fig:coarseiter}-\ref{fig:feet}.


\begin{figure}[tbp]
\centering
\subfigure[$k=0$]{\includegraphics[width=0.4\textwidth]{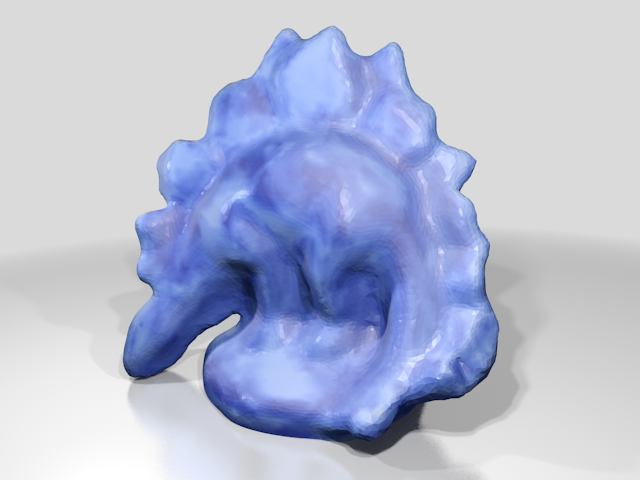}}\hspace{0.15cm}
\subfigure[$k=1$]{\includegraphics[width=0.4\textwidth]{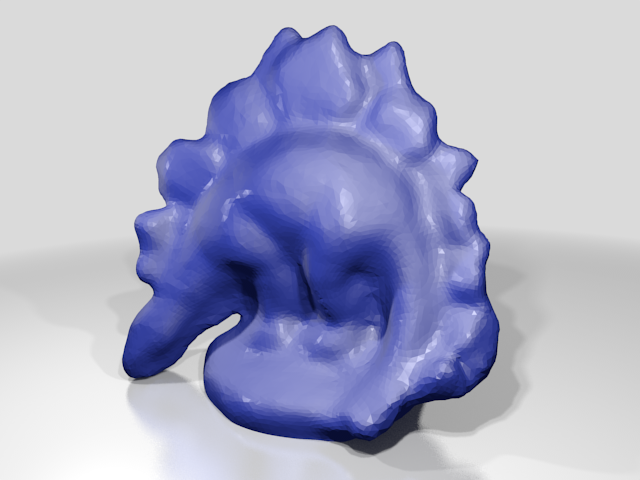}}
\caption{As indicated by the fast decay of the re-projection error distribution, the superlinear convergence rate is maintained for the dino example.}
\end{figure}

\begin{figure}[tbp]
\centering
\subfigure[$k=1$]{\includegraphics[width=0.4\textwidth]{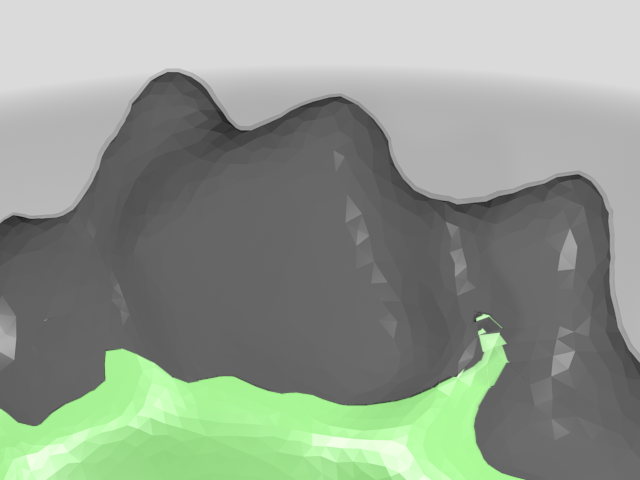}}\hspace{0.15cm}
\subfigure[$k=5$]{\includegraphics[width=0.4\textwidth]{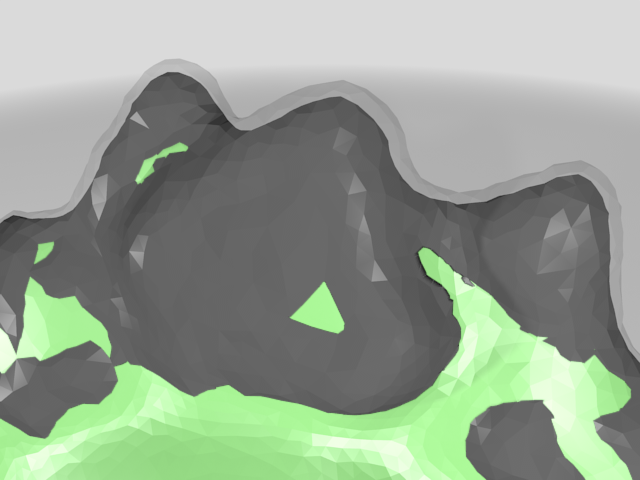}}
\caption{This excerpt of the mesh evolution shown in Fig.~\ref{fig:coarseiter} supports the hypothesis that no shrinking occurs in the LMD method. In fact, the opposite is true: we observe local expansion relative to the green-colored ground truth model. Let us re-emphasize that no contour constraints are imposed.}
\end{figure}

\begin{figure}[tbp]
\centering
\subfigure[$k=0$]{\includegraphics[width=0.4\textwidth]{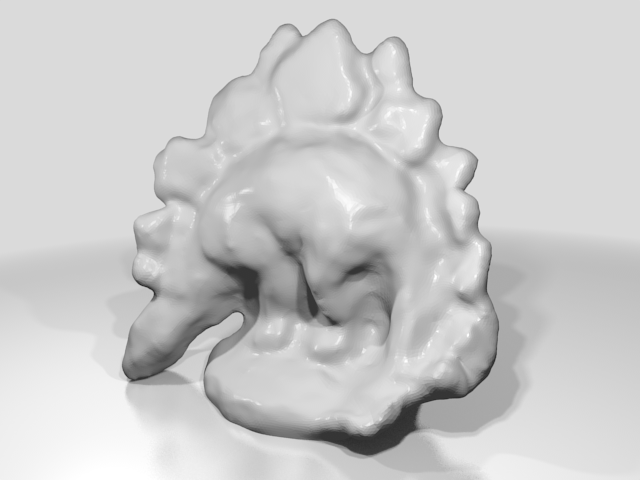}}\hspace{0.15cm}
\subfigure[$k=1$]{\includegraphics[width=0.4\textwidth]{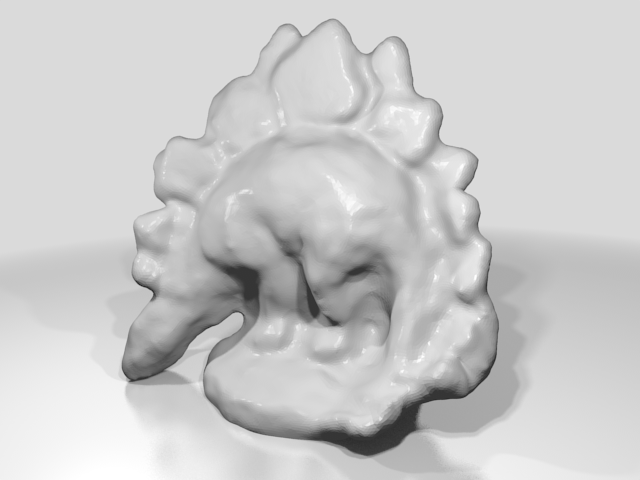}}\\
\subfigure[$k=2$]{\includegraphics[width=0.4\textwidth]{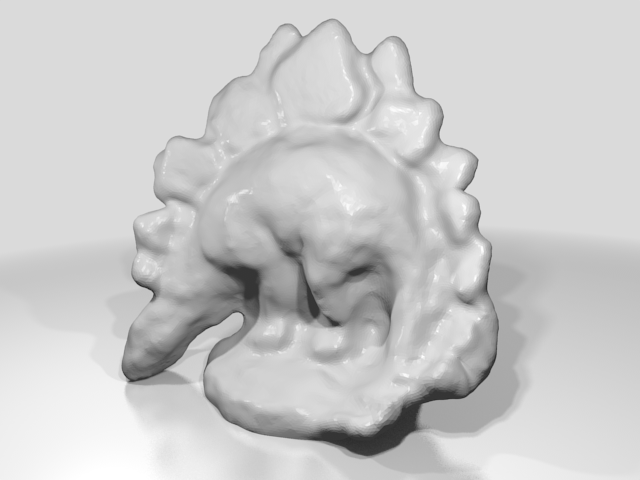}}\hspace{0.15cm}
\subfigure[$k=5$]{\includegraphics[width=0.4\textwidth]{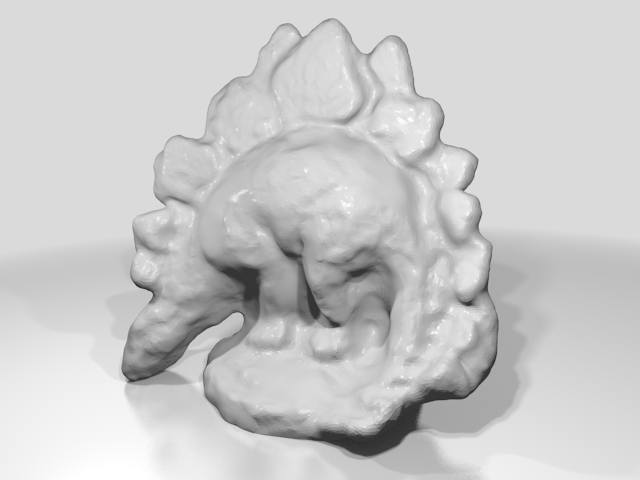}}
\caption{We employ a multiscale mechanism to reduce the computational costs: The further the iteration progresses, the smaller we select the size $\varepsilon$ of the neighborhood in which local image descriptors are computed. Meanwhile, the mesh resolution is increased by Loop subdivision.}
\end{figure}

\subsection{Some remarks on convex programs}

We briefly outline a very simple convex model for multiview stereo reconstruction and discuss its relationship with the present work. Let us assume that the function $p:\RRR\to [0,1]$ -- akin to a probability density -- fulfills  
\begin{subequations}\label{eq:quadraticprogram}
\begin{equation}\label{eq:densityconstraint}
	\int\limits_{\RRR}p\,\mathrm{d}\bm{x}=1.
\end{equation}
It is illustrative to think of $p(\bm{x})$ telling us how likely it is for the surface to pass through the point $\bm{x}$. Since we want the re-projection error $\rho$ to become small where this likelihood is high, we should minimize the ``correlation''
\[
	E_l(p)=\langle\rho,p\rangle_{L^2(\RRR)}=\int\limits_{\RRR}\rho p\,\mathrm{d}\bm{x}
\]
w.r.t. $p$. This naive approach is bound to fail despite the normalization condition~\eqref{eq:densityconstraint} which rules out the optimal $p$ being identically zero. The trivial and possibly non-unique solution would be Dirac's delta distribution $\delta(\bm{x}-\bm{x}^*)$, where $\bm{x}^*$ is a pointwise minimizer of $\rho$. One needs to impose stronger regularity conditions on $p$, e.g., by 
\begin{equation}\label{eq:objective}
	E(p)=\int\limits_{\RRR}\rho p\,\mathrm{d}\bm{x}-\frac{\lambda}{2}\int\limits_{\RRR}\|\nabla p\|^2\,\mathrm{d}\bm{x}.
\end{equation}
\end{subequations}
Altogether, Eq.~\eqref{eq:quadraticprogram} forms a nice convex program: the objective function~\eqref{eq:objective} is a quadric in $p$; Eq.~\eqref{eq:densityconstraint} forces any solution to lie on an infinite-dimensional version of the standard simplex which is a convex set. A solution surface is obtained as the \emph{maximal level set}
\begin{equation}\label{eq:maxlevelset}
	S = \{\bm{x}\in\RRR\;|\; p(\bm{x})=\arg\max_{\RRR}p(\bm{x}),\; p=\arg\min E(p)\}.
\end{equation}
This brings us to the two main objections we have against such a model:

\subsubsection*{Convexity from embeddings}

The key fact to note is that $E_l(p)=\langle\rho,p\rangle_{L^2(\RRR)}$ is a \emph{linear function of $p$}, whereas $E_x(S)$ from Sect.~\ref{subsec:mvs} depends on $S$ \emph{nonlinearly}. Similar applies to the domains of $E_l(p)$ and $E_x(S)$: While shape spaces are manifolds with non-vanishing curvature, $p$ can be taken from a linear space of scalar-valued functions (or a convex subset thereof), whose elements are identified with surfaces at hand of their maximal level set~\eqref{eq:maxlevelset}. Such implicit surface representations provide embeddings of shape spaces into vector spaces but -- due to the extra dimension -- lead to less efficient algorithms than explicit representations. Now, since linearity of $E_l$ carries over to convexity of~\eqref{eq:objective}, the majority of convex models rely on aforementioned embeddings in one way or another. E.g., Kolev et al.~\cite{Kolev2009} use the characteristic function of the volume enclosed by $S$. Their thresholding step corresponds to the selection~\eqref{eq:maxlevelset} of the maximal level set.

\subsubsection*{Visibility} 

Except for very special cases, no analytical formula exists describing the unknown surface let alone its visibility in the given set of views. Since $\rho$ in Eq.~\eqref{eq:objective} crucially depends on visibility, one is forced to operate with numerical approximations. But these are only available through an estimate of the surface itself. Indeed, many state-of-the-art MVS methods such as~\cite{Furukawa2010} can be decomposed into an initialization phase, where  a coarse solution is found by means of sparse image correspondence, and an approximation phase, where the \emph{surface} model is sought that best explains \emph{both}, the results of the first phase \emph{and} the raw image data\footnote{This is where the shape optimization approach presented here could prove valuable.}. In a way, this chicken-and-egg problem defeats the purpose of convex models whose charm is that they are independent of an initial guesses. But as just argued, visibility makes some initial guess indispensable, and even more importantly, the solution will depend on its precise value. 

\bibliographystyle{alpha}

\end{document}